\pdfoutput=1
\documentclass[11pt]{article}
\usepackage{acl} 

\usepackage{times}
\usepackage{latexsym}
\usepackage[T1]{fontenc} % For proper rendering and hyphenation of words containing Latin characters (including in bib files)
\usepackage[utf8]{inputenc}

\usepackage{microtype}
\usepackage{inconsolata}
\usepackage[english]{babel}
\usepackage{amsthm}
 \usepackage{amsmath}
\usepackage{graphicx} 
\usepackage{booktabs}
\usepackage{multirow}
\usepackage{subcaption} 
\usepackage{hyperref}
\usepackage{cleveref}
\usepackage{pgfplots}
\pgfplotsset{compat=1.18}
\usepackage{float}
\usepackage{xcolor}
\usepackage{amssymb}
\usepackage{comment}
\usepackage[colorinlistoftodos]{todonotes}

\newtheorem{researchQuestion}{RQ}

\newcommand{\bz}{{\boldsymbol{z}}}

\newcommand{\vm}[1]{{#1}} % nothing 

\title{On the Representations of Entities in Auto-regressive Large Language Models}

\author{
  Victor Morand$^{1}$ \quad
  Josiane Mothe$^{2}$ \quad
  Benjamin Piwowarski$^{1}$ \\
  \\
  $^{1}$Sorbonne Université, CNRS, \\ 
  Institut des Systèmes Intelligents et de Robotique (ISIR),\\ F-75005 Paris, France \\
  \\
  $^{2}$INSPE, UT2J, Université de Toulouse, IRIT \\ UMR5505 CNRS, F-31400 Toulouse, France \\
  \\
}

\begin{document}

\maketitle

\begin{abstract}
Named entities are fundamental building blocks of knowledge in text, grounding factual information and structuring relationships within language. Despite their importance, it remains unclear how Large Language Models (LLMs) internally represent entities. Prior research has primarily examined explicit relationships, but little is known about entity representations themselves.
We introduce entity mention reconstruction as a novel framework for studying how LLMs encode and manipulate entities. We investigate whether entity mentions can be generated from internal representations, how multi-token entities are encoded beyond last-token embeddings, and whether these representations capture relational knowledge.
Our proposed method, leveraging \textit{task vectors}, allows to consistently generate multi-token mentions from various entity representations derived from the LLMs hidden states. We thus introduce the \textit{Entity Lens}, extending the \textit{logit-lens} to predict multi-token mentions.
Our results bring new evidence that LLMs develop entity-specific mechanisms to represent and manipulate any multi-token entities, including those unseen during training.\footnote{Code is available \href{https://github.com/VictorMorand/EntityRepresentations}{\texttt{here}}, including all the hyper-parameters used in the experiments. }
\end{abstract}

%We show that LLMs store entity representations in a structured manner, enabling consistent reconstruction. We show that multi-token entity representations preserve richer semantic information, even for unseen entities during training,
%and that simple transformations within the representation space can model relational knowledge.
%structures within their internal representation space, rather than treating entities as mere token sequences.
%These insights challenge existing views on knowledge is distributed within LLMs, highlighting the need for representation-based approaches to model interpretability.

\begin{comment}
    TLDR: one sentence describing the paper:
    "This paper adresses the challenge of associating named entities to internal representations in LLMs, we propose a new method that allows to generate the correct mention from well-represented entities."
\end{comment}

\section{Introduction}

%%%%%%%%%%%% 
Despite the remarkable achievements of LLMs across a range of natural language processing tasks, the mechanisms underlying their ability to represent and manipulate knowledge remain opaque, making it challenging to enhance their interpretability and control.
Among the various components of knowledge expression, entities are fundamental building blocks. They serve as anchors for factual information and relationships within text. 
%Unlike general lexical items, entities encapsulate rich, multi-dimensional knowledge, allowing LLMs to ground facts and contextual information in a meaningful way. 
A key challenge is therefore to understand how LLMs internally represent named entities (\textit{e.g.} \texttt{Eiffel tower}) in a given context and how these representations persist, evolve, and encode information across model layers.
Understanding how entity knowledge is stored and retrieved may enable us to better monitor when and why LLMs hallucinate or generate factually inconsistent responses.

Current interpretability research has primarily focused on entity relationships and factual knowledge rather than the internal representation of entities themselves \cite{gevaTransformerFeedForwardLayers2021, gevaDissectingRecallFactual2023, hernandezLinearityRelationDecoding2024, niuWhatDoesKnowledge2023}. 
Many named entities span multiple tokens, yet prior work has largely used single-token probing. 
Our work seeks to fill this gap by exploring how LLMs build unified entity representations from multiple token.
%Specifically, it remains unclear whether entity representations are fully reconstructable from internal activations, or whether knowledge about an entity is distributed across tokens and layers in ways that make direct recovery more difficult. 
%% VM : Too early to talk about reconstruction here IMHO, we convinced the reader that entity representations are important, Now we show the gap in the litterature, and justify why we want to study this association, the reconstruction paradigm naturally comes after we phrased "our primary objective".

\begin{figure}[t]
    \centering
    \hspace{-1cm}
    \includegraphics[scale=.36]{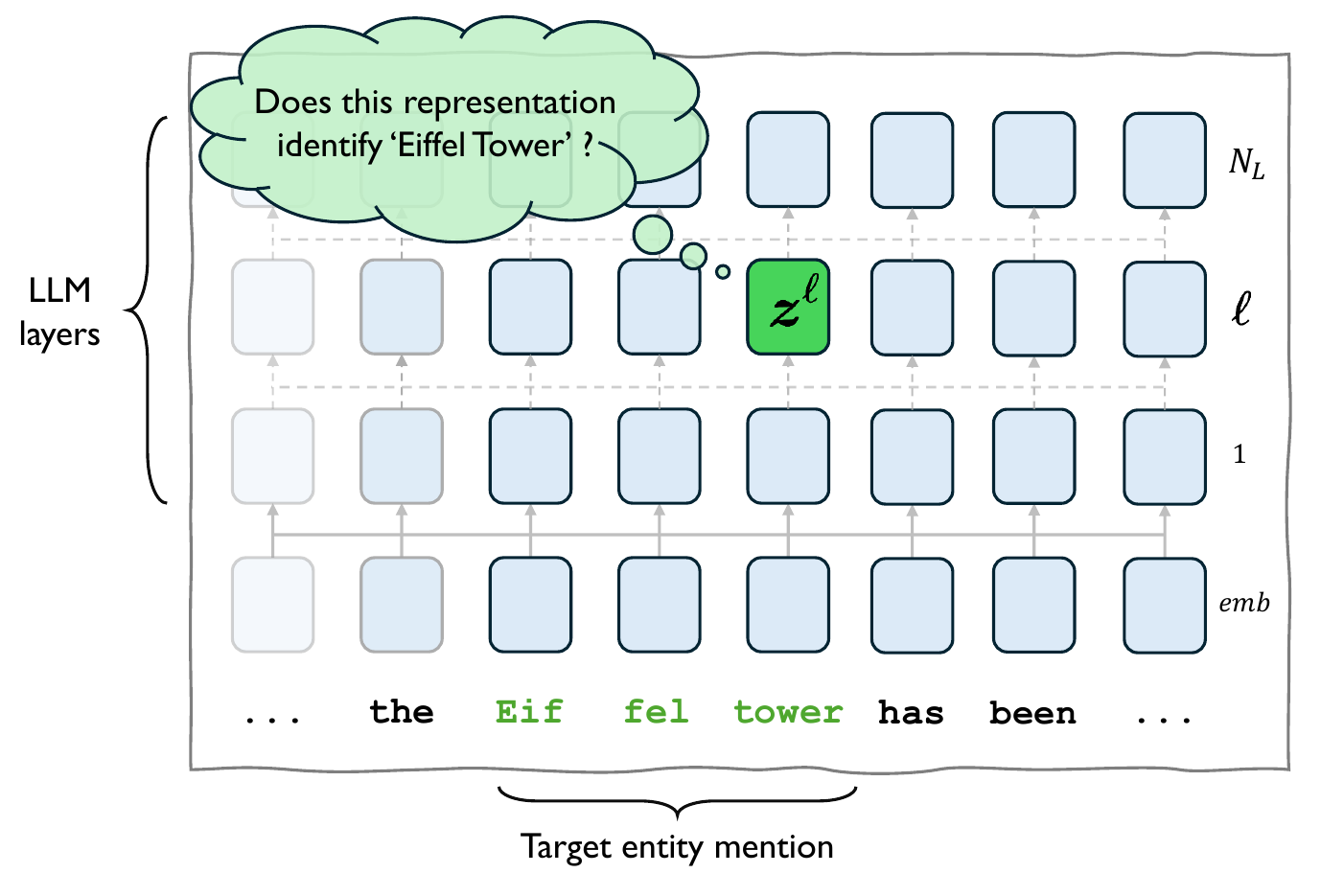}
    \caption{Representation extraction principle:  the representation $\bz^\ell$ (in green) of the last token of a target entity mention - here \textit{tower} for the named entity \textit{Eiffel tower} - is extracted at a given layer $\ell$ of the transformer model.}
    \label{fig:ExtractionMethod}
    \vspace{-0.5cm}
\end{figure}

%For instance, \citet{gevaDissectingRecallFactual2023} found that factual associations, such as subject-attribute mappings, can be traced back to the middle layers of transformer networks. 
%There however remains a gap in understanding how well LLMs capture individual entity meanings beyond simple relational knowledge, especially in the case of multi-token mentions.
%Studying entity reconstruction offers several key insights. If entity representations can be recovered, this suggests that LLMs store knowledge in a structured manner, rather than simply associating facts with token co-occurrences.

%Our work seeks to fill this gap by investigating how LLMs internally represent named entities and exploring whether these representations can be effectively manipulated to decode entity mentions.

%%%%%%%%%%% comment associer les representation aux entités
%Traditional knowledge representation methods, like Word2Vec, demonstrated that structured information could be projected into continuous vector spaces. While LLMs have the potential to extend this concept by encoding dynamic and context-sensitive representations, 

%previous version: I think we need to explain that we choose to measure the quality by retrieving mentions 
%%%%%%%%%%%%%%%%%%%%%%%%%%%%%%%%%%%%%%%%%
%Our central hypothesis is that there exists a representational space within LLMs where entities are meaningfully encoded. We aim to investigate the robustness of these representations and their association with entity mentions. To explore this, we propose the following research questions:
In this work, we posit that LLMs compute layer-agnostic entity representations that can be isolated and manipulated. Our primary objective is to establish a direct association between these internal representations and named entities, focusing  specifically on their mention in text. 
\textit{To evaluate this association, we measure how accurately the corresponding mention can be generated from the representation at hand.} 
%By measuring the fidelity of this reconstruction, we assess how well LLMs preserve entity identity and whether entity-specific knowledge remains accessible across layers.
The core of our methodology is presented in Section~\ref{sec:Background}, addressing the following research question:

\begin{researchQuestion}\label{rq:whichLayer}
How well entity mentions can be decoded from their representation? 
\end{researchQuestion}

%We explore decoding according to several directions. First, our experiment are inline with previous findings that  middle layers better represent entities~\citep{mengLocatingEditingFactual2022}. 
Our results show that LLMs possess specific mechanisms for representing and manipulating entities, allowing them to consistently generate mentions from their internal entity representations. 
We also find that decoding capacity depends more on an entity frequency in the training data than on its token length.

% IN appendix Now -> ref later
%In \Cref{sec:reprConstruction} we analyze where the representation is built inside the LLM. Considering  successive representation and causal analysis, we show there is no clear location where the entity is "completed".

Our next research question challenges the current assumption of using the last token's representation at a given layer. We address it in \Cref{sec:BetterReps}.
\begin{researchQuestion}\label{rq:betterReps}
Can we find better representations than the last token representations from LLMs?
\end{researchQuestion}
%RQ3: \textit{Would combining information from multiple tokens yields better representations?}
%\textit{Is using the last token the best approach for representing multi-token entities?} We explore whether combining information from multiple tokens yields better representations.
%We found that averaging the representations of the entity's tokens as well as cleaning the representation thanks to a linear transformation help in decoding it more accurately. 
As alternatives, we propose averaging the mention token representations and cleaning the entity representation, and show those help the model to decode entity mentions. % We propose two alternative representations for entities that facilitates the generation of their mention.

%Our last research question studies how relations can be associated with simple linear transformations in the representation space.
Our last research question relates to the additional knowledge entity representations capture; it is developed in \Cref{sec:Relation_extraction}:

\begin{researchQuestion}\label{rq:MoreThanMention}
Does the structure of the entity representation space capture (relational) knowledge?
\end{researchQuestion}

%%%%%%%  CONCLUSION. 
By transforming the representation of subject to related object entities with a linear transformation, we show that we can extract more than the sole entity mention from those representations. 

We also introduce a practical application of our method called the \textit{Entity Lens}. It allows to visualize which entity the model is ``thinking'' about at a given layer, extending \textit{logit lens} \cite{nostalgebraistInterpretingGPTLogit2020} to generate multi-token entity mentions, instead of projecting a token’s hidden state onto the vocabulary.
\begin{figure*}[h!] %bt
    \begin{subfigure}[t]{0.49\textwidth}
        \centering
        \includegraphics[scale=.24]{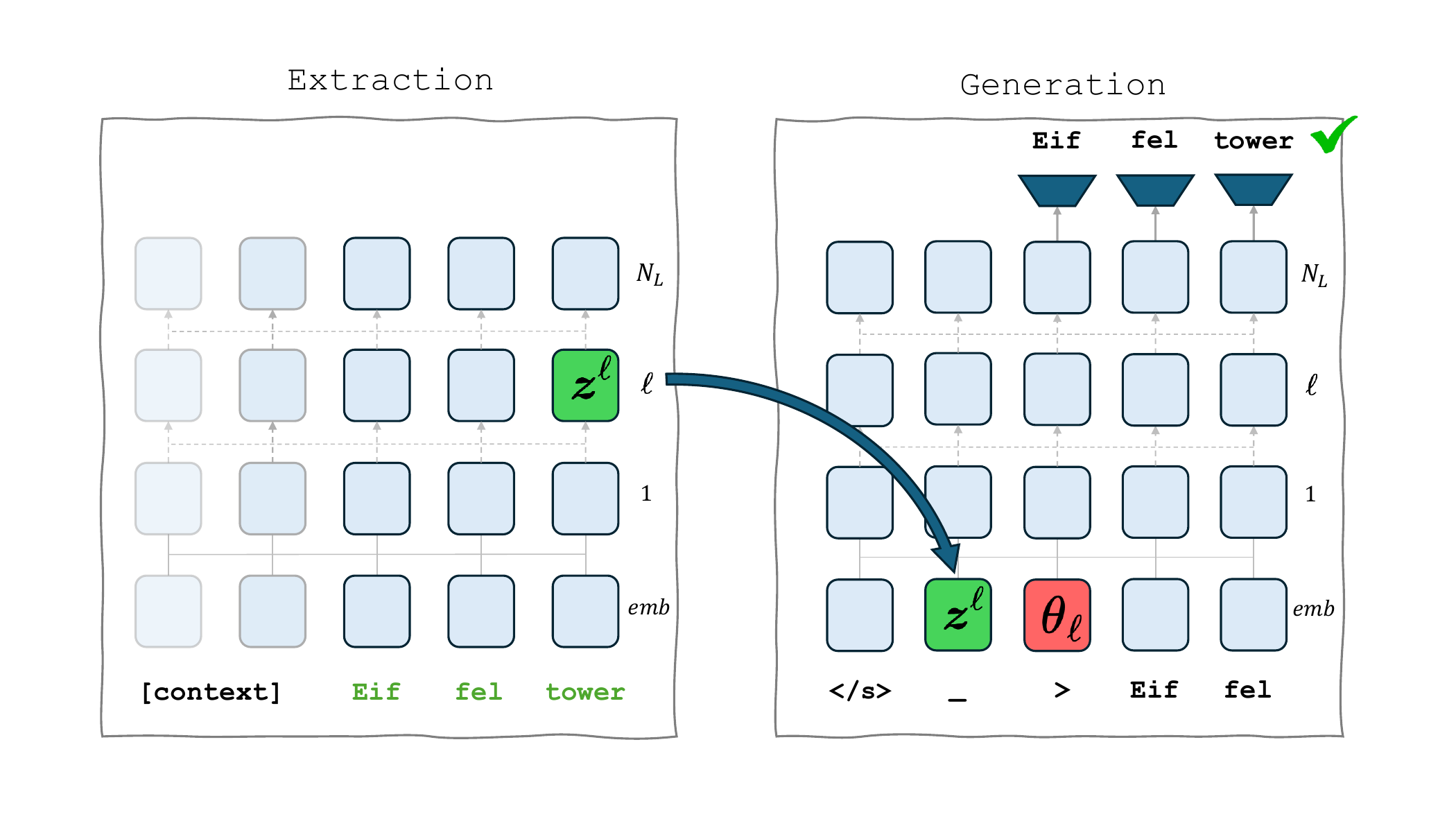}
        \subcaption{ \hyperref[par:UncontextualDecoding]{Uncontextual entity mention decoding}:  The entity representation $\bz^\ell$ at layer $\ell$ is  extracted in context (left, green), its mention is then generated using a learned \textit{task vector} $\theta_\ell$ that prompts the model to generate the mention from $\bz^\ell$ only. }
        \label{fig:schemaNoContext}
    \end{subfigure}
    \hfill
    \begin{subfigure}[t]{0.49\textwidth}
        \centering
        \includegraphics[scale=.26]{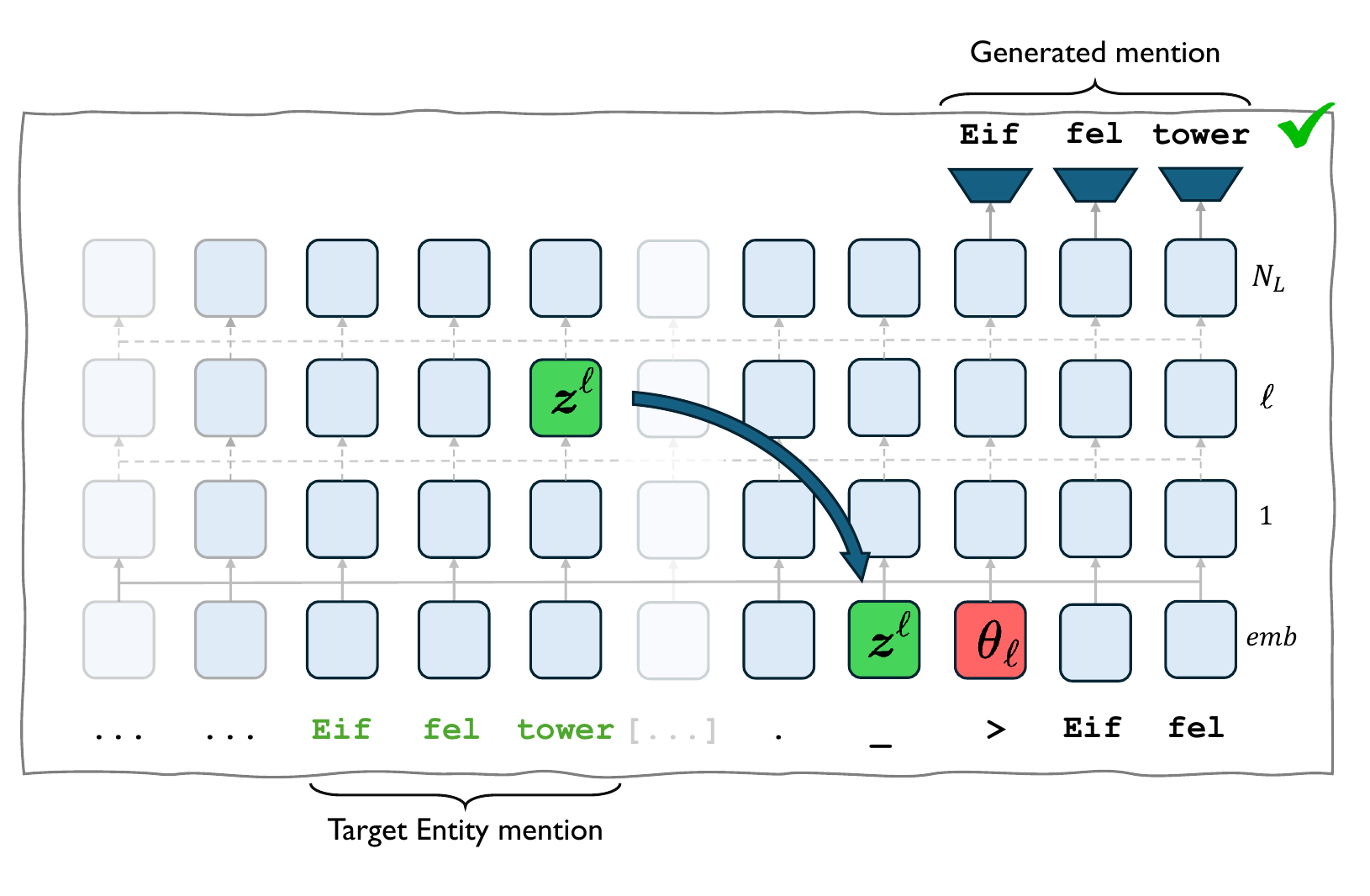}
        \subcaption{\hyperref[par:contextualDecoding]{Contextual entity mention decoding}: Also using a learned  \textit{task vector} $\theta_\ell$, the entity mention is now generated using both the extracted representation $\bz^\ell$ and the surrounding context from which it can be copied. }
        \label{fig:schemaContext}
    \end{subfigure}
    \caption{Our entity mention decoding method, in both uncontextual (left) and contextual (right) senarii.}
    \vspace{-0.3cm}
\end{figure*}

\section{Related Work} \label{sec:RelatedWork}

%attention, c'est élaboré avec du ChatGPT

% The representation and manipulation of knowledge within language models has evolved from traditional knowledge bases and static embeddings to sophisticated techniques for probing and understanding how large language models (LLMs) internally encode information. 

 % \textbf{Traditional knowledge representation in NLP} used structured knowledge bases.%like DBpedia, which stored facts as triples (subject, predicate, object) \cite{auer2007dbpedia}.

% ---- Entity and knowledge representation in early works

The question of knowledge representation has been studied very early in the development of neural network approaches for language modeling. Word2Vec~\cite{mikolov2013efficient}  shows that relationships between words (e.g., singular to plural, capital to country) could be represented as translations between word representations. This %in turn 
motivated the development of knowledge base representations where entities were linked by geometrical transformation -- e.g., TransE~\cite{Bordes2013TranslatingEF} that explicitly uses the translation observed in Word2Vec.

% While useful for querying, these approaches struggled with scalability and coverage. The shift to distributed representations began with Word2Vec \cite{mikolov2013efficient}, which captured semantic similarities in dense vectors.
With the development of contextual word -- and then token -- embeddings, such as ELMo \cite{peters2019knowledge} and BERT \cite{devlin2018bert}, the question of how knowledge is encoded and how entities are represented was temporarily put aside. 

Then, early work~\cite{petroni2019language} showed that pre-trained LLMs can act as "knowledge bases" by retrieving factual information through prompt-based queries.
To understand how this retrieval occurs in LLMs, 
\citet{gevaTransformerFeedForwardLayers2021,mengLocatingEditingFactual2022} hypothesized that feed-forward layers act as key-value memories, storing and retrieving factual associations during inference. Even if this idea  has been challenged~\cite{niuWhatDoesKnowledge2023},
\citet{hernandezLinearityRelationDecoding2024} showed that, within LLMs, simple relations can often be approximated by a linear mapping of the subject to the object entity representation at a middle decoder layer -- echoing  \cite{mikolov2013efficient} in the context of LLMs. This suggests that  LLMs operate, at least partially, within a structured entity representation space where entities can be manipulated.
Recent research employing sparse auto-encoders supports this hypothesis, with \citet{ferrando2025iknowentityknowledge} identifying a feature within this entity representation space that quantifies how much the model "knows" about an entity.

% Going further, some works have been looking at how relations are decoded within a LLM~ , the authors examined the linearity of relation decoding in LLMs, suggesting that different layers contribute differently to encoding entity-relationship information.
% This aligns with work by \cite{niuWhatDoesKnowledge2023}, who investigated the "knowledge neuron" hypothesis, positing that specific neurons in the network may be responsible for encoding distinct types of factual knowledge. 

This raises the question of how entities themselves are represented. Research suggests that earlier layers capture surface-level details, while deeper layers encode more abstract and task-specific features \cite{jawahar2019does, voita2019bottom, gevaDissectingRecallFactual2023}. While entities are processed across all layers, their representations—especially for multi-token entities—may not always be coherent or robust. %This layered approach to information refinement suggests that while entities are represented throughout the network, their representations may not always be coherent or robust, particularly for multi-token entities.

%Probing methods --> before logit lens
While probing methods are used to determine whether specific types of information (e.g., syntactic structure or factual content) can be extracted from the model's representations \cite{conneau2018you, tenney2019bert}, these probes are not fine-grained and do not allow to understand how knowledge is processed and represented.
To gain some insight on how LLM processes knowledge, a specific type of probe, called `logit attribution' or `logit lens' has been developed~\cite{nostalgebraistInterpretingGPTLogit2020, yuCharacterizingMechanismsFactual2023, dalvi2019one}. With logit lens, the LLM hidden states are projected onto the vocabulary using the unembedding matrix, allowing to identify which tokens would be predicted at a given layer of the LLM.

% In a follow-up study, \cite{gevaDissectingRecallFactual2023} showed that subject-attribute mappings are often captured in the lower layers of auto-regressive language models, while deeper layers refine this information for tasks such as question answering.
Representing multi-token entities, like `\texttt{Eif|fel|\_tower}', presents unique challenges for LLMs. Their internal representation may fail to capture the entity’s full meaning consistently. Using the logit lens can help, but is insufficient to properly associate representations with the entire multi-token mention.
This issue is further highlighted by  \cite{liu2021does}, who found that tokenization granularity can significantly affect the quality of representations for multi-word expressions. \textit{In this contribution, we extend logit lens by proposing the Entity lens that decodes full entity mentions from internal representations.}
To address these challenges, researchers have explored strategies like attention-based aggregation \cite{clark2019does}, contextual subword pooling \cite{schick2020s}, and token-level masking \cite{joshi2020spanbert}.  More recent methods such as Patchscopes \cite{ghandehariounPatchscopesUnifyingFramework2024} and SelfIE \cite{chenSelfIESelfInterpretationLarge2024} extends the limitations of the logit lens to produce multi-token explanations from an extracted representation. In contrast with generating general explanations, our work focuses on entity mention, allowing the use of quantitative metrics such as exact match. Our work specifically studies the problem of how to represent and manipulate multi-token entities.

\section{Retrieving entity mentions from LLM representations}\label{sec:MentionDecoding}

\subsection{Methodology}\label{sec:Background}

In transformer-based language models \cite{vaswaniAttentionAllYou2017},  text is tokenized into a sequence of tokens $(t_1, \ldots , t_n) \in \mathcal V^n$, with $\mathcal V$ the vocabulary used by the tokenizer. These tokens are embedded into a sequence of initial \textit{representations} $(\bz_1^0, \ldots , \bz_n^0) \in \mathbb R^d$, with $d$ the model's representation space dimension. These representations are sequentially passed through the transformer layers: each layer $\ell \in \{1,\ldots,N_L\}$ generates a new series of {representations} $(\bz_1^\ell, \ldots, \bz_n^\ell) \in \mathbb{R}^d$, building on the representations from the preceding layer. 

%To measure whether an internal representation of entities exists,  we extract the hidden representation $\bz_\ell$ corresponding to an entity mention at a specific layer $\ell$ of the  model. Our experiments are designed to investigate how well these extracted representations can be bound to the entity mention from which they were extracted. We evaluate exact matches, meaning that the entity must be fully reconstructed to be considered correct (see Figure~\ref{fig:ExtractionMethod}). 

% on explique le but: décoder une mention complète à partir d'une rep extraite du dernier token
\paragraph{Decoding}\label{par:Decoding} In this contribution, we aim at associating a given representation $\bz \in \mathbb{R}^d$ at hand with a named entity by trying to generate the mention from which the representation was extracted.
Following the literature~\cite{mengLocatingEditingFactual2022,gevaDissectingRecallFactual2023}, $\bz^\ell$ is extracted from the last token of the entity mention at layer $\ell$ (See  \Cref{fig:ExtractionMethod}). \Cref{sec:BetterReps} extends our experiments to other representations.
To generate a mention for the representation $\bz^\ell$, we insert it bypassing the embedding layer and \textit{prompt} the model to reconstruct the corresponding entity mention. This is done using a \textit{soft prompt} or \textit{embedding vector}, $\theta_\ell \in \mathbb{R}^d$, optimized for the task. This method is known as \textit{Prompt Tuning} \cite{lesterPowerScaleParameterEfficient2021}.
Because this vector is functional rather than semantic, we refer to $\theta_\ell$ as a \textit{task vector}, inspired by \citet{hendelInContextLearningCreates2023}.
%$\theta_\ell$ is optimized for each transformer layer separately and the model decodes the entity mention using only the extracted representation.
%The decoding process differs depending on whether context is provided. In both scenariis, we assess the quality of the entity representation $\bz^\ell \in \mathbb{R}^d$ by generating the original mention from which it was extracted. The quality of reconstruction gives insight into how well the representation captures the entity.

\paragraph{Motivation of using Task Vectors} Prompt tuning is a parameter-efficient method to prompt the model to perform a variety of tasks \cite{lesterPowerScaleParameterEfficient2021}. It is however not the only method that can be used to generate text. Since our focus is on interpretability while keeping the LLM unchanged, fine-tuning is excluded. Probes have also been explored in this context (see for instance \citet{palFutureLensAnticipating2023}), but they do not easily allow to decode an arbitrary-lenght sequences of tokens, nor to prompt the model to retrieve a mention from a context. To the best of our knowledge, our task vector approach is  the only method that can address all these constraints and challenges of entity reconstruction. Our results furthermore show that this method  performs well for the considered task.

% dans le cas contextuel
\paragraph{Uncontextualized Decoding} \label{par:UncontextualDecoding}
%Following previous work \cite{mengLocatingEditingFactual2022, gevaDissectingRecallFactual2023}, in our study, $\bz^\ell$ is extracted from the final token of the entity mention at layer $\ell$ (see Figure~\ref{fig:schemaNoContext}).  
In this  setting, the model's only input is the representation $\bz^\ell$ along with the task vector $\theta_\ell$. The goal is to generate the entire entity mention.
This allows us to measure how much information about the entity mention is retained in $\bz^\ell$. This setting is illustrated  \Cref{fig:schemaNoContext}.

\paragraph{Contextualized Decoding.} \label{par:contextualDecoding}
%In contrast to the previous setting, i
In contextualized decoding, we include the textual context from which the representation was extracted before prompting the transformer with the task vector $\theta_\ell$.  \Cref{fig:schemaContext} provides an overview of this methodology. 
In this setting, the model can just copy the mention from the context. This is however not a trivial task as the model must first identify the correct span within the context,\textit{ i.e.} essentially performing Named Entity Recognition (NER).

%mtn que le lecteur a compris (espérons) ce qu'on fait, on précise que l'on est au courant du pb des entités ambigües, mais que notre paradigme reste justifié

%\begin{comment}    
% on précise la métrique utilisée + on justifie que l'on a bien conscience du pb de la désambiguation, qui ne fait que limiter notre performance sans atteindre au bien fondé du problème que l'on pose.
% je ne sais pas vous voulez insister sur une limite qui je ne suis pas sure soit vraiement l'objet. En plus vos exemples sont les mêmes que ceux de la section 6 qui montre que le modèle fait des liens. Cela amène de la confusion il me semble
% vrai pour les liens avec entity lens, à modifier si ce paragrpahe convient en essence. 
% Je suis d'accord que ce n'est pas vrmt l'objet mais mais en même temps c'est un des points demandés par le meta Reviewer, d'ou le fait que j'en ai initialement fait un paragraphe dédié. 
% c'est une interrogation légitime du lecteur qui comprend ce que l'on fait, autant la désamorcer directement non ? (cela a finit comme une critique dure d'un des reviewers qui disait que notre pb était mal posé)
% possible aussi de le remettre dans les limitations, où il se trouvait initialement.

\paragraph{Evaluation} 
% We consider a vector $\bz\in \mathbb R^d$ to be a good representation of an entity \textit{if} the model can generate the correct mention from it.
\vm{In both settings, our setup allows to generate a complete mention from a given entity representation $\bz\in \mathbb R^d$. We can then evaluate the reconstruction performance using the exact match metric (EM) between generated mentions and the original mentions from which the representations were extracted.
Entity mentions are however known to be ambiguous, and a more general problem would consider all possible mentions for a given entity (\textit{e.g.} \texttt{NYC} or \texttt{New York City}). The metric used in our \hyperref[par:UncontextualDecoding]{uncontextual} setup only measures the ability to reconstruct the specific mention from which the representation was extracted in the context, without accounting for possible additional ambiguous mentions. 
\textit{Although well posed, the results are therefore only a lower bound for the more general problem}. For instance, \texttt{New York City} could be considered as an accurate generation from the representation of the mention \texttt{NYC}; our evaluation metric, however, considers it a failure.  
This limitation does not apply in the \hyperref[par:contextualDecoding]{contextual} setup, where the model is instructed to copy the mention directly from the provided context.
}

%%% version de Josiane
%This allows the model to use both the task vector $\theta_\ell$ and the surrounding text to reconstruct the entity. The addition of context helps the model better identify the correct tokens associated with the entity, especially when the entity is ambiguous or consists of multiple tokens (See \Cref{fig:schemaContext}).  
%By including context, we can examine how much the additional information improves the model’s ability to generate the correct mention. The same process of extracting representations and generating mentions is followed.

\paragraph{Task Vector Training} In both settings, for each layer $\ell$ of the model, the task vector $\theta_\ell$ is learned by maximizing the log-likelihood of generating the entity mention, given both the representation and $\theta_\ell$. 
We use cross-entropy, the standard loss for language modeling tasks.
%The parameters are optimized using Adam \cite{kingma2017adammethodstochasticoptimization}.
%The reason we train a separate task vector for each layer is to capture how entity information is distributed across different layers in the transformer model. 
\vm{More implementation details and code can be found in the reproducibility statement, in \Cref{sec:Reproductibility}.}
Training a task vector for each layer allows us to evaluate which layers provide the most accurate entity representations.
% Earlier layers may encode surface-level details of the entity, while deeper layers are expected to capture more abstract or context-dependent representations. 

\subsection{Dataset and Experiment Setup}\label{subsec:Dataset}
\begin{figure}[htb]
    \centering
        % This file was created with tikzplotlib v0.10.1.
\begin{tikzpicture}
\tikzstyle{every node}=[font=\small]

\definecolor{darkgray176}{RGB}{176,176,176}
\definecolor{darkblue}{RGB}{70,90,230}

\begin{axis}[
tick align=outside,
tick pos=left,
x grid style={darkgray176},
width = 6.5cm,
height = 4cm,
xlabel={Number of tokens in entity mention},
xmajorgrids,
xmin=-0.195, xmax=16.195,
xtick style={color=black},
y grid style={darkgray176},
ylabel={Counts},
ymajorgrids,
ymin=0, ymax=1810.6,
ytick style={color=black}
]
\draw[draw=none,fill=darkblue] (axis cs:0.55,0) rectangle (axis cs:1.45,704);
\draw[draw=none,fill=darkblue] (axis cs:1.55,0) rectangle (axis cs:2.45,1646);
\draw[draw=none,fill=darkblue] (axis cs:2.55,0) rectangle (axis cs:3.45,1338);
\draw[draw=none,fill=darkblue] (axis cs:3.55,0) rectangle (axis cs:4.45,854);
\draw[draw=none,fill=darkblue] (axis cs:4.55,0) rectangle (axis cs:5.45,428);
\draw[draw=none,fill=darkblue] (axis cs:5.55,0) rectangle (axis cs:6.45,248);
\draw[draw=none,fill=darkblue] (axis cs:6.55,0) rectangle (axis cs:7.45,105);
\draw[draw=none,fill=darkblue] (axis cs:7.55,0) rectangle (axis cs:8.45,52);
\draw[draw=none,fill=darkblue] (axis cs:8.55,0) rectangle (axis cs:9.45,8);
\draw[draw=none,fill=darkblue] (axis cs:9.55,0) rectangle (axis cs:10.45,3);
\draw[draw=none,fill=darkblue] (axis cs:10.55,0) rectangle (axis cs:11.45,2);
\draw[draw=none,fill=darkblue] (axis cs:14.55,0) rectangle (axis cs:15.45,1);
\addplot [black, dashed]
table {%
3 0
3 1810.6
};
\draw (axis cs:3.3,1646) node[
  scale=0.5,
  anchor=base west,
  text=black,
  rotate=0.0
]{Median: 3.0 tokens};
\end{axis}

\begin{axis}[
axis y line=right,
tick align=outside,
width = 6.5cm,
height = 4cm,
x grid style={darkgray176},
xmin=-0.195, xmax=16.195,
xtick pos=left,
xtick style={color=black},
y grid style={darkgray176},
ymin=0, ymax=33.5980701428837,
ytick pos=right,
ytick style={color=black},
yticklabel style={anchor=west},
ytick={0,10,20,30},
yticklabels={0\%,10\%,20\%,30\%}
]
\addplot [semithick, red, opacity=0.0, dashed, mark=*, mark size=3, mark options={solid}]
table {%
1 13.0636481722026
2 30.5437001298942
3 24.8283540545556
4 15.8470959361663
5 7.94210428650956
6 4.6019669697532
7 1.94841343477454
8 0.964928558174058
9 0.148450547411394
10 0.0556689552792726
11 0.0371126368528484
15 0.0185563184264242
};
\end{axis}

\end{tikzpicture}
    \vspace{-1cm}
    \caption{Distribution of the number of tokens for named entity mentions in the test set (\textsc{Pythia} tokenizer). 87\% of them are tokenized with two or more tokens. The dashed line is the median token count (3).}
    \label{fig:LabelLengthsDistrib}
    \vspace{-.4cm}
\end{figure}
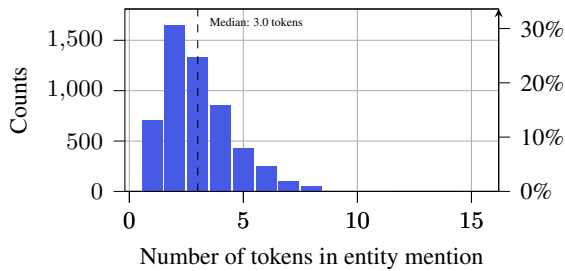

\paragraph{Data} \vm{Our experiment involves extracting the representation of an entity mention within its context, and then trying to reconstruct the entity mention from this representation. This requires a dataset containing sentences with labeled spans of entity mentions: in other words, a NER dataset, for which we ignore the entity categories.} We use the CoNLL-2003 dataset \cite{sang2003introductionconll2003sharedtask}, a widely used NER benchmark. It provides a diverse set of named entities across various types, lengths and frequencies, making it well-suited for studying entity representations.
In the CoNLL-2003 dataset and using the Pythia tokenizer, most entity mentions are tokenized into 2 or 3 tokens, while some span up to 8-12 tokens (see ~\Cref{fig:LabelLengthsDistrib}). This diversity in token length allows us to explore the robustness of the models in handling both short and long multi-token entities, a key aspect of this study. \vm{More details on the dataset statistics can be found in \Cref{tab:DataStats}.}

\paragraph{Models}
We focus on decoder-only architectures, used in the vast majority of recent models, and known for their superior performance.
We experiment with different models from two families. First, the \textsc{Pythia} family \cite{bidermanPythiaSuiteAnalyzing2023}, which includes several models of various sizes, trained with the same data and close settings. This allows the study of the impact of architecture parameters on performance, especially the model size.
We also use the recent \textsc{Phi} family, a set of models trained with the latest techniques and curated data \cite{textbooks2phi1_5}. We use the available non-instruct versions, namely \textsc{Phi-1.5} \cite{textbooks2phi1_5} and \textsc{Phi-2} \cite{bubeckPhi2SurprisingPower2023}, allegedly trained on identical text data, with the particularity that knowledge from \textsc{Phi-1.5} has been distilled into \textsc{Phi-2}. We also use \textsc{Phi-3}\footnote{more specifically \href{https://huggingface.co/microsoft/Phi-3-mini-4k-instruct}{\texttt{Phi-3-mini-4k-instruct}}.} \cite{abdinPhi3TechnicalReport2024}, the next iteration of the \textsc{Phi} family which  follows the \textsc{Llama-2} architecture, and has the particularity of having a significantly smaller vocabulary size (32k tokens compared to 50k tokens for all other models considered in this work) and of being instruction fine-tuned -- which is the only version available.
We utilize models ranging from 140m to 7B parameters, which offers a fair range for studying the impact of model size, while limiting resource consumption. The architecture parameters of all the models we experimented with are detailed in appendix (\Cref{table:modelParams}).

\subsection{Results}    
%To address \Cref{rq:whichLayer}, we use multiple models and decode entities from their representation at various layers. We also break down the results according to the number of tokens an entity is composed of.  We consider both uncontextual and contextual setups. 
%Without context, decoding obviously heavily depends on the frequency of the entity mention in the training data. For less frequent entities, we show that we can use in context decoding to measure the quality of the representation.

%\subsection{Layer perspective}

%In previous work, \citet{morrisTextEmbeddingsReveal2023} managed to reconstruct exactly 92\% of 32-token texts. %, from 768-dimensional representations computed with text-embedding models (\textit{e.g} GTR-T5-Base, \citet{ni2021largedualencodersgeneralizable}).  This suggests that transformer models are able to compress information about many tokens into a single vector. Here we study the specific case of named entities.

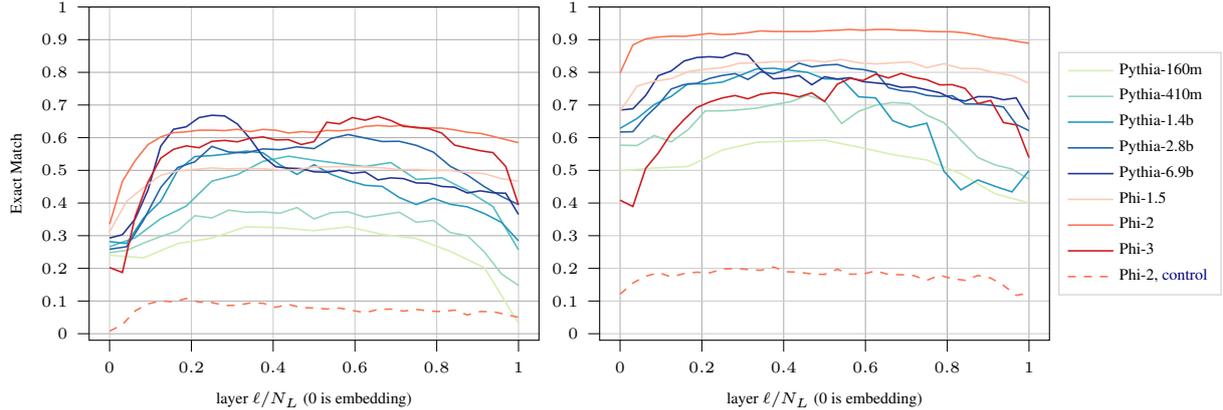
\begin{figure*}[ht]
    %\hspace{-.5cm}
    \begin{subfigure}[t]{0.43\textwidth}
        \centering
        % This file was created with tikzplotlib v0.10.1.
\begin{tikzpicture}
\tikzstyle{every node}=[font=\tiny]

\definecolor{coral25111684}{RGB}{251,116,84}
\definecolor{darkgray176}{RGB}{176,176,176}
\definecolor{darkslateblue3654149}{RGB}{36,54,149}
\definecolor{firebrick2072831}{RGB}{207,28,31}
\definecolor{lightgray204}{RGB}{204,204,204}
\definecolor{lightgreen151214184}{RGB}{151,214,184}
\definecolor{lightseagreen36152192}{RGB}{36,152,192}
\definecolor{mediumturquoise81188193}{RGB}{81,188,193}
\definecolor{palegoldenrod214239178}{RGB}{214,239,178}
\definecolor{peachpuff252201180}{RGB}{252,201,180}
\definecolor{steelblue33100171}{RGB}{33,100,171}

\begin{axis}[
legend cell align={left},
legend to name=empty, % This removes the legend
width = 7.5cm,    % Set the desired width
height= 6cm,    % Set the desired heigh
tick align=outside,
tick pos=left,
x grid style={lightgray204},
xlabel={layer $\ell / N_L$ (0 is embedding)},
xmajorgrids,
xmin=-0.05, xmax=1.05,
xtick style={color=black},
y grid style={darkgray176},
ylabel={Exact Match},
ytick={0,0.1,...,1},
ymajorgrids,
ymin=-0.02, ymax=1,
ytick style={color=black}
]

\addplot [semithick, palegoldenrod214239178, opacity=1, solid]
table {%
0 0.239933197253665
0.0833333333333333 0.232325106698831
0.166666666666667 0.275932455000928
0.25 0.292447578400445
0.333333333333333 0.327333457042123
0.416666666666667 0.324364446093895
0.5 0.315271850064947
0.583333333333333 0.327333457042123
0.666666666666667 0.303952495824828
0.75 0.29133419929486
0.833333333333333 0.251994804230841
0.916666666666667 0.20207830766376
1 0.0315457413249211
};
\addlegendentry{Pythia-160m}
\addplot [semithick, lightgreen151214184, opacity=1, solid]
table {%
0 0.248097977361292
0.0416666666666667 0.254963815179068
0.0833333333333333 0.277416960475042
0.125 0.29745778437558
0.166666666666667 0.315086286880683
0.208333333333333 0.361105956578215
0.25 0.354425681944702
0.291666666666667 0.378177769530525
0.333333333333333 0.371497494897012
0.375 0.372796437186862
0.416666666666667 0.368899610317313
0.458333333333333 0.386156986453888
0.5 0.351085544627946
0.541666666666667 0.370384115791427
0.583333333333333 0.373167563555391
0.625 0.356281313787345
0.666666666666667 0.363889404342179
0.708333333333333 0.371868621265541
0.75 0.341065132677677
0.791666666666667 0.346817591389868
0.833333333333333 0.309333828168491
0.875 0.300055668955279
0.916666666666667 0.250139172388198
0.958333333333333 0.182965299684543
1 0.147893857858601
};
\addlegendentry{Pythia-410m}
\addplot [semithick, mediumturquoise81188193, opacity=1, solid]
table {%
0 0.266283169419187
0.0625 0.293560957506031
0.125 0.352384486917796
0.1875 0.390796066060494
0.25 0.466876971608833
0.3125 0.486917795509371
0.375 0.52755613286324
0.4375 0.5433290035257
0.5 0.531081833364261
0.5625 0.520875858229727
0.625 0.511412135832251
0.6875 0.524030432362219
0.75 0.472258303952496
0.8125 0.47726850992763
0.875 0.437929114863611
0.9375 0.38931156058638
1 0.256633883837447
};
\addlegendentry{Pythia-1b}
\addplot [semithick, lightseagreen36152192, opacity=1, solid]
table {%
0 0.281870476897384
0.0416666666666667 0.276118018185192
0.0833333333333333 0.354611245128966
0.125 0.405269994433104
0.166666666666667 0.500278344776396
0.208333333333333 0.542215624420115
0.25 0.544999072184079
0.291666666666667 0.550380404527742
0.333333333333333 0.558916311003897
0.375 0.553720541844498
0.416666666666667 0.516793468175914
0.458333333333333 0.487660048246428
0.5 0.517350157728707
0.541666666666667 0.487845611430692
0.583333333333333 0.468918166635739
0.625 0.454258675078864
0.666666666666667 0.450918537762108
0.708333333333333 0.416960475041752
0.75 0.395249582482835
0.791666666666667 0.414548153646317
0.833333333333333 0.396177398404157
0.875 0.388198181480794
0.916666666666667 0.367786231211728
0.958333333333333 0.340694006309148
1 0.284097235108554
};
\addlegendentry{Pythia-1.4b}
\addplot [semithick, steelblue33100171, opacity=1, solid]
table {%
0 0.258489515680089
0.0416666666666667 0.266283169419187
0.0833333333333333 0.348487660048246
0.125 0.447392837261087
0.166666666666667 0.508814251252552
0.208333333333333 0.526071627389126
0.25 0.573390239376508
0.291666666666667 0.555947300055669
0.333333333333333 0.553534978660234
0.375 0.56559658563741
0.416666666666667 0.563184264241974
0.458333333333333 0.571905733902394
0.5 0.566524401558731
0.541666666666667 0.597513453330859
0.583333333333333 0.609575060308035
0.625 0.598812395620709
0.666666666666667 0.588235294117647
0.708333333333333 0.588977546854704
0.75 0.576544813509
0.791666666666667 0.555761736871405
0.833333333333333 0.512339951753572
0.875 0.485618853219521
0.916666666666667 0.449805158656523
0.958333333333333 0.420300612358508
1 0.395064019298571
};
\addlegendentry{Pythia-2.8b}
\addplot [semithick, darkslateblue3654149, opacity=1, solid]
table {%
0 0.292818704768974
0.03125 0.303024679903507
0.0625 0.35127110781221
0.09375 0.439784746706254
0.125 0.5739469289293
0.15625 0.625719057339024
0.1875 0.633327147893858
0.21875 0.661532751902023
0.25 0.668584152904064
0.28125 0.666542957877157
0.3125 0.641863054370013
0.34375 0.596771200593802
0.375 0.542586750788643
0.40625 0.511226572647987
0.4375 0.506773056225645
0.46875 0.505288550751531
0.5 0.495453701985526
0.53125 0.494154759695676
0.5625 0.500092781592132
0.59375 0.49081462237892
0.625 0.495082575616998
0.65625 0.470773798478382
0.6875 0.477082946743366
0.71875 0.474670625347931
0.75 0.46149563926517
0.78125 0.460567823343849
0.8125 0.450732974577844
0.84375 0.448506216366673
0.875 0.430692150677306
0.90625 0.437372425310818
0.9375 0.431434403414363
0.96875 0.429949897940249
1 0.365188346632028
};
\addlegendentry{Pythia-6.9b}

%%%%%% END Pythia models

\addplot [semithick, peachpuff252201180, opacity=1, solid,]
table {%
0 0.312117275932455
0.0416666666666667 0.405269994433104
0.0833333333333333 0.447578400445352
0.125 0.484876600482464
0.166666666666667 0.49823714974949
0.208333333333333 0.502690666171832
0.25 0.507515308962702
0.291666666666667 0.504546298014474
0.333333333333333 0.505845240304324
0.375 0.505102987567267
0.416666666666667 0.500463907960661
0.458333333333333 0.501948413434775
0.5 0.510855446279458
0.541666666666667 0.510855446279458
0.583333333333333 0.511226572647987
0.625 0.51048431991093
0.666666666666667 0.50788643533123
0.708333333333333 0.501948413434775
0.75 0.502876229356096
0.791666666666667 0.499350528855075
0.833333333333333 0.499721655223604
0.875 0.492855817405827
0.916666666666667 0.489886806457599
0.958333333333333 0.474485062163667
1 0.466691408424569
};
\addlegendentry{Phi-1.5}
\addplot [semithick, coral25111684, opacity=1, solid]
table {%
0 0.335498237149749
0.03125 0.46632028205604
0.0625 0.529782891074411
0.09375 0.578771571720171
0.125 0.602152532937465
0.15625 0.615327519020226
0.1875 0.617925403599926
0.21875 0.62293560957506
0.25 0.62293560957506
0.28125 0.621636667285211
0.3125 0.625719057339024
0.34375 0.618482093152719
0.375 0.625347930970495
0.40625 0.62293560957506
0.4375 0.614214139914641
0.46875 0.618482093152719
0.5 0.616069771757283
0.53125 0.622193356838003
0.5625 0.621079977732418
0.59375 0.624791241417703
0.625 0.634440526999443
0.65625 0.637223974763407
0.6875 0.634626090183708
0.71875 0.638337353868992
0.75 0.629430321024309
0.78125 0.629987010577101
0.8125 0.629244757840045
0.84375 0.623863425496382
0.875 0.616626461310076
0.90625 0.61328632399332
0.9375 0.604008164780108
0.96875 0.594173316014103
1 0.584709593616626
};
\addlegendentry{Phi-2}
\addplot [semithick, firebrick2072831, opacity=1, solid]
table {%
0 0.202449434032288
0.03125 0.18723325292262
0.0625 0.358879198367044
0.09375 0.463907960660605
0.125 0.53720541844498
0.15625 0.56429764334756
0.1875 0.574874744850622
0.21875 0.569307849322694
0.25 0.58879198367044
0.28125 0.591575431434403
0.3125 0.587307478196326
0.34375 0.598070142883652
0.375 0.602338096121729
0.40625 0.593245500092782
0.4375 0.593987752829839
0.46875 0.579142698088699
0.5 0.587864167749119
0.53125 0.647244386713676
0.5625 0.642234180738541
0.59375 0.661532751902023
0.625 0.65355353497866
0.65625 0.664872889218779
0.6875 0.652811282241603
0.71875 0.634440526999443
0.75 0.631842642419744
0.78125 0.627389125997402
0.8125 0.614028576730377
0.84375 0.577101503061793
0.875 0.568380033401373
0.90625 0.560029690109482
0.9375 0.554277231397291
0.96875 0.511968825385044
1 0.394507329745778
};
\addlegendentry{Phi-3}

\addplot [semithick, coral25111684, opacity=1, dashed]
table {%
0.0  0.007785289899610736
0.03125  0.02771504824471361
0.0625  0.06994290375203915
0.09375  0.09109477124183006
0.125  0.10052910052910052
0.15625  0.09752099979512395
0.1875  0.10772066352652059
0.21875  0.09681372549019608
0.25  0.09555964804583589
0.28125  0.08627370997348562
0.3125  0.08704536166734778
0.34375  0.09405537459283388
0.375  0.09317531671434409
0.40625  0.08081632653061224
0.4375  0.0895675343308055
0.46875  0.07978831671076735
0.5  0.07879159012043274
0.53125  0.07554469558134799
0.5625  0.07830709466366796
0.59375  0.07113738279657562
0.625  0.06442405708460755
0.65625  0.07430087773014901
0.6875  0.07538304392236976
0.71875  0.06902218570254724
0.75  0.07430213464696224
0.78125  0.06948145111703218
0.8125  0.06766612641815235
0.84375  0.07225964482547459
0.875  0.05728848114169215
0.90625  0.0678069639584606
0.9375  0.06744091279543603
0.96875  0.05878750765462339
1.0  0.05027590435315757
};
\addlegendentry{baseline (Phi-2)}

\end{axis}

\end{tikzpicture}
        \vspace{-1cm}
        \subcaption{\hyperref[par:UncontextualDecoding]{Uncontextual mention decoding} results by layer.}
        \label{figsub:UncontextResults}
    \end{subfigure}
    \hfill
    \begin{subfigure}[t]{0.58\textwidth}
        \centering
        % This file was created with tikzplotlib v0.10.1.
\begin{tikzpicture}
\tikzstyle{every node}=[font=\tiny]

\definecolor{coral25111684}{RGB}{251,116,84}
\definecolor{darkgray176}{RGB}{176,176,176}
\definecolor{darkslateblue3654149}{RGB}{36,54,149}
\definecolor{lightgray204}{RGB}{204,204,204}
\definecolor{lightgreen151214184}{RGB}{151,214,184}
\definecolor{lightseagreen36152192}{RGB}{36,152,192}
\definecolor{palegoldenrod214239178}{RGB}{214,239,178}
\definecolor{peachpuff252201180}{RGB}{252,201,180}
\definecolor{steelblue33100171}{RGB}{33,100,171}
\definecolor{firebrick2072831}{RGB}{207,28,31}

\begin{axis}[
legend cell align={left},
legend style={
  fill opacity=0.6,
  draw opacity=1,
  text opacity=1,
  at={(1.02,0.5)},  % Position the legend outside the plot
  anchor=west,      % Anchor the legend to the west (right side)
  draw=lightgray204,
  legend columns=1  % Set the legend to one column
},
width = 7.5cm,    % Set the desired width
height= 6cm,    % Set the desired heigh
tick align=outside,
tick pos=left,
x grid style={darkgray176},
xlabel={layer $\ell / N_L$ (0 is embedding)},
xmajorgrids,
xmin=-0.05, xmax=1.05,
xtick style={color=black},
y grid style={lightgray204},
ytick={0,0.1,...,1},
ymajorgrids,
ymin=-0.02, ymax=1,
ytick style={color=black}
]

\addplot [semithick, palegoldenrod214239178, opacity=1, solid]
table {%
0 0.498422712933754
0.0833333333333333 0.507329745778438
0.166666666666667 0.511783262200779
0.25 0.562442011504917
0.333333333333333 0.585822972722212
0.416666666666667 0.589348673223232
0.5 0.59231768417146
0.583333333333333 0.571163481165337
0.666666666666667 0.548339209500835
0.75 0.531638522917053
0.833333333333333 0.489701243273335
0.916666666666667 0.428465392466135
1 0.400445351642234
};
\addlegendentry{Pythia-160m}
\addplot [semithick, lightgreen151214184, opacity=1, solid]
table {%
0 0.576915939877528
0.0416666666666667 0.575988123956207
0.0833333333333333 0.606606049359807
0.125 0.586936351827797
0.166666666666667 0.624976804601967
0.208333333333333 0.681944702171089
0.25 0.681944702171089
0.291666666666667 0.685284839487846
0.333333333333333 0.69029504546298
0.375 0.700872146966042
0.416666666666667 0.70996474299499
0.458333333333333 0.730005566895528
0.5 0.712748190758953
0.541666666666667 0.643347559844127
0.583333333333333 0.68064575988124
0.625 0.699387641491928
0.666666666666667 0.707552421599555
0.708333333333333 0.704397847467063
0.75 0.668955279272592
0.791666666666667 0.646131007608091
0.833333333333333 0.59231768417146
0.875 0.539432176656151
0.916666666666667 0.51660790499165
0.958333333333333 0.504175171645945
1 0.472629430321024
};
\addlegendentry{Pythia-410m}
\addplot [semithick, lightseagreen36152192, opacity=1, solid]
table {%
0 0.628502505102988
0.0416666666666667 0.658934867322323
0.0833333333333333 0.70124327333457
0.125 0.726479866394507
0.166666666666667 0.765448135089998
0.208333333333333 0.764334755984413
0.25 0.770087214696604
0.291666666666667 0.789756912228614
0.333333333333333 0.810354425681945
0.375 0.812581183893116
0.416666666666667 0.806457598812396
0.458333333333333 0.799962887363147
0.5 0.779922063462609
0.541666666666667 0.779179810725552
0.583333333333333 0.725180924104658
0.625 0.721655223603637
0.666666666666667 0.652069029504546
0.708333333333333 0.631842642419744
0.75 0.644275375765448
0.791666666666667 0.497680460196697
0.833333333333333 0.440155873074782
0.875 0.470588235294118
0.916666666666667 0.451289664130636
0.958333333333333 0.434032287994062
1 0.499536092039339
};
\addlegendentry{Pythia-1.4b}
\addplot [semithick, steelblue33100171, opacity=1, solid]
table {%
0 0.617368714047133
0.03125 0.618296529968454
0.0625 0.664130636481722
0.09375 0.701428836518835
0.125 0.748933011690481
0.15625 0.76247912414177
0.1875 0.766190387827055
0.21875 0.781406568936723
0.25 0.790499164965671
0.28125 0.796066060493598
0.3125 0.778066431619967
0.34375 0.801261829652997
0.375 0.79569493412507
0.40625 0.807014288365188
0.4375 0.828168491371312
0.46875 0.816849137131193
0.5 0.822230469474856
0.53125 0.824086101317499
0.5625 0.812210057524587
0.59375 0.808684357023567
0.625 0.80051957691594
0.65625 0.75282983856003
0.6875 0.74410836889961
0.71875 0.740025978845797
0.75 0.729448877342735
0.78125 0.725923176841715
0.8125 0.72833549823715
0.84375 0.703098905177213
0.875 0.702913341992949
0.90625 0.699758767860456
0.9375 0.694006309148265
0.96875 0.640564112080163
1 0.621636667285211
};
\addlegendentry{Pythia-2.8b}
\addplot [semithick, darkslateblue3654149, opacity=1, solid]
table {%
0 0.684542586750789
0.03125 0.688439413620338
0.0625 0.728892187789942
0.09375 0.789942475412878
0.125 0.804230840601225
0.15625 0.834477639636296
0.1875 0.845425867507886
0.21875 0.847838188903322
0.25 0.845054741139358
0.28125 0.859343106327705
0.3125 0.852848394878456
0.34375 0.808869920207831
0.375 0.779922063462609
0.40625 0.783262200779365
0.4375 0.790313601781407
0.46875 0.761922434588978
0.5 0.786416774911857
0.53125 0.779550937094081
0.5625 0.783633327147894
0.59375 0.771757283354982
0.625 0.767674893301169
0.65625 0.76377806643162
0.6875 0.754128780849879
0.71875 0.755242159955465
0.75 0.746335127110781
0.78125 0.737799220634626
0.8125 0.719428465392466
0.84375 0.712005938021896
0.875 0.72573761365745
0.90625 0.725180924104658
0.9375 0.71608832807571
0.96875 0.721840786787901
1 0.655223603637038
};
\addlegendentry{Pythia-6.9b}

\addplot [semithick, peachpuff252201180, opacity=1, solid]
table {%
0 0.681016886249768
0.0416666666666667 0.757840044535164
0.0833333333333333 0.772685099276304
0.125 0.780478753015402
0.166666666666667 0.802189645574318
0.208333333333333 0.809797736129152
0.25 0.814436815735758
0.291666666666667 0.827797365002783
0.333333333333333 0.829096307292633
0.375 0.832436444609389
0.416666666666667 0.832065318240861
0.458333333333333 0.836889961031731
0.5 0.831879755056597
0.541666666666667 0.839487845611431
0.583333333333333 0.829281870476897
0.625 0.825756169975877
0.666666666666667 0.828910744108369
0.708333333333333 0.831879755056597
0.75 0.814251252551494
0.791666666666667 0.826498422712934
0.833333333333333 0.812024494340323
0.875 0.81146780478753
0.916666666666667 0.799035071441826
0.958333333333333 0.790313601781407
1 0.766375951011319
};
\addlegendentry{Phi-1.5}
\addplot [semithick, coral25111684, opacity=1, solid]
table {%
0   0.798014473928373
0.03125     0.883806519453207
0.0625  0.902331910682254
0.09375     0.908053442197068
0.125   0.910836889961032
0.15625     0.910001855631843
0.1875  0.91439351765943
0.21875     0.919094451660791
0.25    0.915352260778128
0.28125     0.917176965423393
0.3125  0.921444918661471
0.34375     0.926826251005134
0.375   0.925001546359869
0.40625     0.924970619162492
0.4375  0.92475412878085
0.46875     0.924939691965114
0.5     0.927259231768417
0.53125     0.928867446032041
0.5625  0.931310694624853
0.59375     0.928589101255644
0.625   0.931310694624853
0.65625     0.931063277045834
0.6875  0.928805591637286
0.71875     0.928527246860889
0.75    0.925403599925775
0.78125     0.924537638399208
0.8125  0.92416651203068
0.84375     0.920548029937527
0.875   0.914115172883033
0.90625     0.906197810354426
0.9375  0.902331910682254
0.96875     0.894785674522175
1   0.889095070204738
};
\addlegendentry{Phi-2}

\addplot [semithick, firebrick2072831,  opacity=1, solid]
table {%
0 0.408795694934125
0.03125 0.38931156058638
0.0625 0.506030803488588
0.09375 0.553534978660234
0.125 0.613100760809056
0.15625 0.655223603637038
0.1875 0.692336240489887
0.21875 0.709222490257933
0.25 0.721655223603637
0.28125 0.729077750974207
0.3125 0.718871775839673
0.34375 0.732974577843756
0.375 0.73798478381889
0.40625 0.733902393765077
0.4375 0.724995360920393
0.46875 0.737242531081833
0.5 0.710892558916311
0.53125 0.763592503247356
0.5625 0.781777695305251
0.59375 0.77472629430321
0.625 0.79439599183522
0.65625 0.782891074410837
0.6875 0.796808313230655
0.71875 0.784746706253479
0.75 0.777509742067174
0.78125 0.761736871404713
0.8125 0.762293560957506
0.84375 0.750788643533123
0.875 0.704212284282798
0.90625 0.714047133048803
0.9375 0.646316570792355
0.96875 0.639079606606049
1 0.538875487103359
};
\addlegendentry{Phi-3}

\addplot [semithick, coral25111684, opacity=1, dashed]
table {%
0.0  0.12097103223174215
0.03125  0.15507364975450083
0.0625  0.17583316295236148
0.09375  0.18670756646216768
0.125  0.1738599348534202
0.15625  0.18387293830177154
0.1875  0.18818737270875763
0.21875  0.18612244897959185
0.25  0.19771708112515288
0.28125  0.19983719983719983
0.3125  0.19657771440211855
0.34375  0.1939840392879067
0.375  0.20427263479145474
0.40625  0.18966218966218967
0.4375  0.18803418803418803
0.46875  0.1829543126408523
0.5  0.18128298822574096
0.53125  0.19726083401471792
0.5625  0.18207739307535642
0.59375  0.1825010150223305
0.625  0.1920408163265306
0.65625  0.18143031288094272
0.6875  0.1805527123848516
0.71875  0.17780040733197555
0.75  0.16327788046826863
0.78125  0.17903752039151713
0.8125  0.16812895327484187
0.84375  0.16331505179768435
0.875  0.17822990844354017
0.90625  0.17038705713700594
0.9375  0.14805725971370143
0.96875  0.11745513866231648
1.0  0.124185667752443
};
\addlegendentry{Phi-2, \hyperref[par:baseline]{control}}

\end{axis}

\end{tikzpicture}
        \vspace{-1cm}
        \subcaption{\hyperref[par:contextualDecoding]{Contextual mention decoding} results by layer.}
        \label{figsub:ContextResults}
    \end{subfigure}
    \caption{
     Context improves decoding. Better performances are obtained on representations extracted in middle layers. Curves present the rate of exact match (on $y$-axis)   after training a task vector on representations extracted at layer $\ell$ ($x$-axis is normalized layer $\ell/N_L$, as model have a different number of layers $N_L$).}
    \label{fig:MentionDecoding}
    \vspace{-0.5cm}
\end{figure*}

%% GENERAL RESULT : It WORKS ! Only afterwards do we consider layer and token perspectives
The results in both settings are presented in \Cref{figsub:UncontextResults} (uncontextual decoding) and \Cref{figsub:ContextResults} (contextual).
Without access to any context, some models (\textsc{Pythia-6.9b}, \textsc{Phi-2} and \textsc{Phi-3}) are able to decode \textit{exactly} the whole mention of up to 65\% of the named entities from the test set. Upon analysis of the results, we observe that failed samples typically exhibit high semantic similarity with the original mention (see \Cref{tab:uncontextualized_samples} in appendix for examples of failures).
Model size unsurprisingly improves performance, supporting a well-known property that larger models memorize more knowledge, including named entities.\footnote{Best performance as a function of model size is reported in appendix, \Cref{fig:AggResultsSizes}.}
\Cref{figsub:UncontextResults} also shows that representations from the middle layers achieve better performance in mention decoding, corroborating findings reported in \cite{mengLocatingEditingFactual2022, mengMassEditingMemoryTransformer2023}.

We unsurprisingly achieve significantly better generation results in the \hyperref[par:contextualDecoding]{contextual setting} where the model can copy the mention from the given context (\Cref{figsub:ContextResults}). Near-optimal performance is reached: 93\% EM when generating entity mentions using representations of their last tokens with \textsc{Phi-2} (layer 20) -- it was only 64\% in the uncontextual setup.
%These results however show that LLMs do not in general store the whole entity mention in representations of its last token. 
These results reveal a key new insight: LLMs do not, in general, store the entire entity mention within the representation of its last token. This challenges the common assumption that the final token’s representation encapsulates the full entity meaning \cite{mengLocatingEditingFactual2022, gevaDissectingRecallFactual2023}, and suggests that alternative encoding approaches may be needed to better capture multi-token entity representations.

Unlike sentence embedding models -- which have been shown to encode almost all tokens from the input sentence \cite{morrisTextEmbeddingsReveal2023} -- auto-regressive language models are not trained to encode tokens explicitly in their internal representations. Instead, if an entity is already mentioned in the context, the model can retrieve it thanks to the attention mechanism and simply copy it, as shown in various mechanistic interpretability studies \cite{wangInterpretabilityWildCircuit2022, mcdougallCopySuppressionComprehensively2023}. 

%Entity mentions are better generated when the model has access to the context (See  \Cref{figsub:ContextResults}). In the \hyperref[par:contextualDecoding]{contextual setting}, the model can just copy the mention from the context.
%When adding the context, 
%Known entities (frequent in the LLM training set) are well predicted, but the model struggles with less common ones (entity mentions that appear less frequently), despite indications that it can retrieve relevant features (\textit{e.g} generating an Italian-sounding name for an Italian personality, see in appendix, \Cref{tab:uncontextualized_samples}). 
%\subsection{Entity size and frequency}

\paragraph{Multi-token entities and frequency}\label{subsec:Length_frequence}

\begin{figure*}
    \centering
    \includegraphics[width=\linewidth]{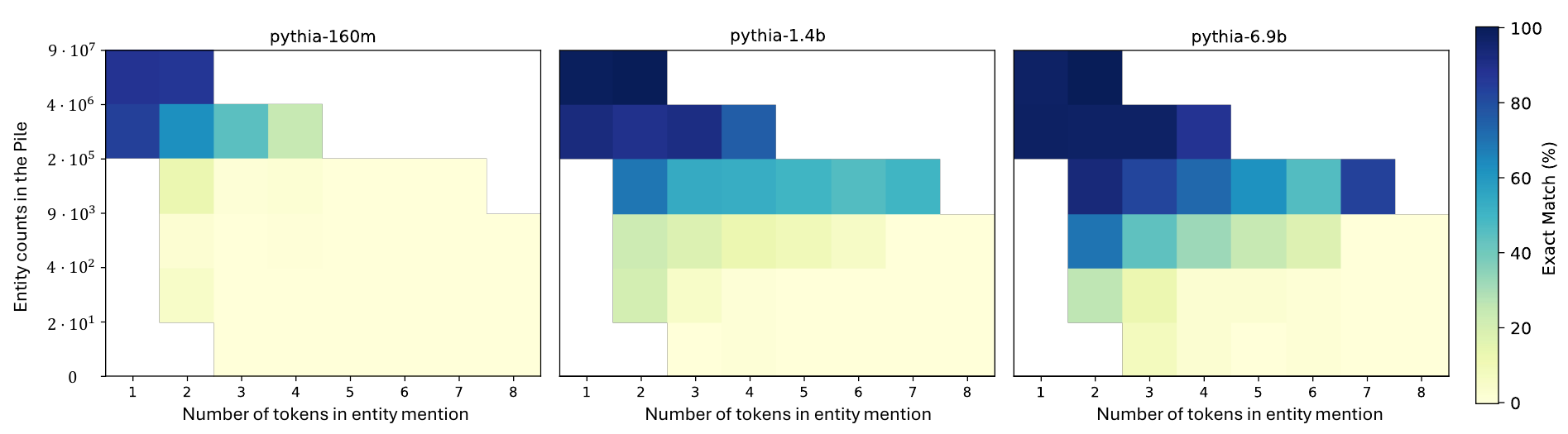}
    \vspace{-.4cm}
    \caption{ \textbf{Uncontextual} mention generation performance is higher for more frequent entities. Performance is analyzed by entity length and mention frequency in \textit{the Pile} \cite{gao2020pile800gbdatasetdiverse}. For each model, we chose the layer with best exact match on the \hyperref[subsec:Dataset]{test set}. Empty cells indicate fewer than five samples. See \Cref{sec:perfNtokenFreq} for full results across models and settings.}
    \label{fig:PerfNtoken_freq_pythias_only3}
    \vspace{-.4cm}
\end{figure*}

Decoding the one-token entity \verb|France| from its embedding or representation at any layer $\bz^\ell$ is much easier than decoding \verb|Mand-el-bro-t| from the sole representation of token \verb|'t'|. 
To further analyze how the number of tokens affects decoding performance, we split the test set based on tokenized mention length. 
Then, following the intuition that entity mentions frequent in the LLM training set should be easier to generate than rare ones,
%~\footnote{The fact the model fails for rare entities is not a limitation as LLMs can copy the mention from context in their regular operating conditions.}, 
we also split the test set into quantiles based on mention n-gram frequency in \textit{the Pile} \cite{gao2020pile800gbdatasetdiverse}.\footnote{$n$-gram frequencies are obtained using the API from \cite{liuInfinigramScalingUnbounded2024}.}
We evaluate both settings across all models. In both cases, entity frequency -- and not the number of tokens -- is the main factor for accurate entity mention decoding (though the two are naturally  correlated, see \Cref{fig:PerfNtoken_freq_pythias_only3} for results on the three \textsc{Pythia} models in the uncontextual setting).

\paragraph{Baseline - Decoding any sequence}\label{par:baseline}
We setup a control experiment to confirm that the unveiled mention decoding ability is specific to entities rather than a general capability of LLMs to decode prior tokens. Using the same methodology as in \Cref{sec:MentionDecoding}, we replace all entity mentions in our original \hyperref[subsec:Dataset]{dataset} with randomly sampled sequences of three\footnote{Three is the median entity length in our data, see \Cref{fig:LabelLengthsDistrib}. We also show in \Cref{subsec:Allbaselines} the generation results for arbitrary sequences of one and two tokens.}  tokens from the data, constraining the last token to be the end of a word for fairer comparison.
We then train new task vectors for each model layer, so they instruct the LLM to \textit{decode the 3-token sequence from the final-token's representation}. The results are presented in both settings in \Cref{fig:MentionDecoding}, (control) restricted to the best-performing model (\textsc{Phi-2}) for clarity.
In the \hyperref[par:UncontextualDecoding]{uncontextual setup}, this baseline reaches only 9\% Exact Match (EM), compared to 65\% when decoding entity mentions, strongly supporting our claim that LLMs detect and represent entities in a specific manner. In the \hyperref[par:contextualDecoding]{contextual setup}, the task seems trivial: The model must only find the extracted token in the context and copy it along with the two preceding tokens. However, results are surprisingly poor, with a maximum 19\% EM (layer 9 of Phi-2, compared to 93\% EM with 3-token entity mentions in the original contextual experiment). This further supports the idea that, unlike counting tokens, manipulating entity mentions --and potentially other meaningful units -- is a natural task for which LLMs develop specialized circuits.

\paragraph{Conclusion}
\vm{Overall, our results in \Cref{sec:MentionDecoding} bring strong evidence that LLMs develop specific mechanisms for representing and manipulating entities.
In both settings, with only a single learned task vector, we successfully  prompt the model to generate the  correct entity mention from its representation.}
For named entities that are frequent in the LLM training set, the last token representation at middle layers is enough to retrieve their whole mention, just as if the latter was part of the vocabulary (\Cref{fig:PerfNtoken_freq_pythias_only3}). For less common entities, LLMs however do not store the whole mention in its last token representation (even if they could, see the supplementary experiment in \Cref{sec:TrainingRepresentations}). Still, when provided with the context, LLMs can very robustly detect and copy them, achieving near optimal performance in our setup (see \Cref{figsub:ContextResults}).
These results also demonstrate that representations are layer-agnostic since the LLM is still able to process them when injected at the embedding level, using only a simple task vector. \vm{Additional experiments investigating how those representations are built are detailed in \Cref{sec:reprConstruction}. We also confirm that task vectors generalize across different settings and layers (see \Cref{subsec:TV_generalization}), further supporting our claim that they activate to a given extent the same specific circuits within LLMs.}

%This representation can be in turn robustly associated with an entity mention, generalizing the \textit{logit-lens} to multi-token predictions. 

\section{Obtaining better representations}\label{sec:BetterReps}
In this section, we question the optimality of using the last token representation of an entity. We extend our initial experiments by using alternative representations for named entity mentions.
For experiments, we chose to focus on \textsc{Phi-2}, because it is newer compared to \textsc{Pythia}, has a reasonable number of parameters allowing fast inference and got the best results in the first set of experiments.

\paragraph{Average representations}\label{par:average}

The most common way to extract representations of sentences with language models is to \textit{average} the representations of all their tokens \cite{jurafsky2009speech}. As superpositions are the building blocks of LLMs \cite{elhageToyModelsSuperposition2022}, this seems a natural choice.
Formally, for a given named entity mention $e = (t_{e_1},...,t_{e_2})$, we compute its average representation at layer $\ell$ over the mention tokens, $\bar \bz^\ell$ as $\bar \bz^\ell = \sum_{i=e_1}^{e_2} \frac{\bz^\ell_i}{e_2-e_1+1}$. 

\paragraph{Training a linear layer to \textit{clean} the entity}\label{par:cleaned}

Representations are highly dependent on the context \cite{ethayarajhHowContextualAre2019}. In other words, for a given entity mention, the representations extracted from the inner layers of transformer models are very likely to contain, along with the representation of the entity,  
a lot of noise that comes from the context. Passing the extracted representation $\bz$ into a linear layer that would clean or reinforce some relevant features might produce entity representations of better quality, in the sense that the model could better decode the entity mention from them.
For both \hyperref[par:UncontextualDecoding]{uncontextual} and \hyperref[par:contextualDecoding]{contextual} settings, we therefore train a linear model with parameters $(W,b)$ that is applied to the extracted representation $\bz$ to obtain a \textit{ cleaned} representation $ \mathcal C_\bz  = W  \bz + b \in \mathbb R^d$.

\begin{figure}[htbp]
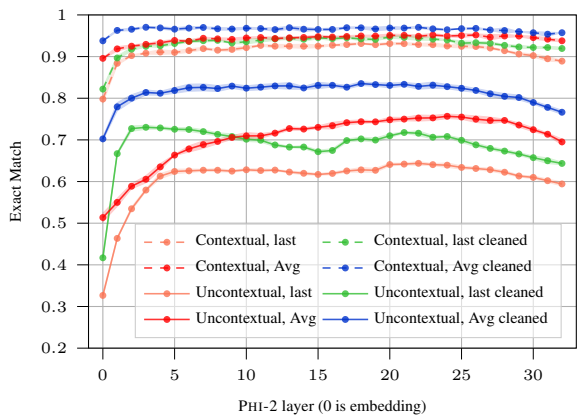

    \centering
    \include{BaselineAvgComp_CoNLL2003_phi-2}%
    \vspace{-1.1cm}
    \caption{Entity mention generation with different representations. Representations are extracted as described in \Cref{sec:Background}(last), by averaging (avg) or cleaning (clean) mention representations, as detailed \Cref{par:average}. Each experiment was conducted 5 times to get an estimate of the variance. (not clearly visible since it is quite small)}
    \vspace{-.5cm}
    \label{fig:BaselineAvgComp}
\end{figure}

%\Cref{fig:BaselineAvgComp} shows the results of both settings when using these new representations. 
%For exhaustiveness, we also combine cleaning with averaging representations.
\paragraph{Results} 
Overall, we see in \Cref{fig:BaselineAvgComp} that in both  \hyperref[par:UncontextualDecoding]{uncontextual} and  \hyperref[par:contextualDecoding]{contextual} setups, the average representation of the tokens from a named entity mention allows to better decode it than the sole representation of the last token (See \Cref{fig:BaselineAvgComp} comparing the red and orange curves).
The gain compared to representations of the last token is aligned with the conclusion of our uncontextual mention generation experiment (see \Cref{figsub:UncontextResults}), suggesting that the representation of the last token in auto-regressive LLMs does not \textit{in general} encode all the tokens of the entity mention.

In the embedding layer (layer $0$ in \Cref{fig:BaselineAvgComp}) representations $\bar \bz^0$ are just the average of embeddings. For uncontextual decoding, performance jumps to 51\% compared to the 33\% of exact matches obtained when generating only from the last token's embedding (solid  red and orange curves) -- this holds to a lesser extent for contextual decoding too. If the model can disentangle all the tokens from their superposed representation in $\bar \bz^0$, it still has to figure out their order to retrieve the original mention.\footnote{For instance, ``ITALY'' is decoded as ``YALIT''} Upper layers representations may therefore encode a notion of relative token position.
The fact that last token representations from middle layers yield better results than the average of the token embeddings (up to 64\% vs 51\%) however shows that there is more than only token superposition in the representation of an entity mention.

Transforming the representation with a linear model before inserting it at the embedding layer also improves the reconstruction performance in both settings (Green and blue curves in  \Cref{fig:BaselineAvgComp}). 
The generation performance gain means that removing or boosting some relevant features -- probably associated with non-entity representation subspaces, \textit{common to all entity representation}, helps the model in generating the correct entity mention. 

Overall, this demonstrates that using the sole last token representation at a given layer may not be the best choice to represent an entity. Further work is needed though to understand precisely the effect of the linear transformation and averaging.

\section{Generalizing relation decoding and logit lens}\label{sec:Relation_extraction}

\paragraph{Relation Extraction} Here, we study a complementary question on whether the discussed entity representations can be manipulated (\Cref{rq:MoreThanMention}). Recent work on the explainability of knowledge manipulation in transformers \cite{mengLocatingEditingFactual2022, mengMassEditingMemoryTransformer2023, gevaDissectingRecallFactual2023, hernandezLinearityRelationDecoding2024, gottesmanEstimatingKnowledgeLarge2024} support the hypothesis that, for a relation $r$ linking subject and object entities $s$ and $o$, the representation of the object $\bz_o$ can be extracted from the representation of the subject $\bz_s$.

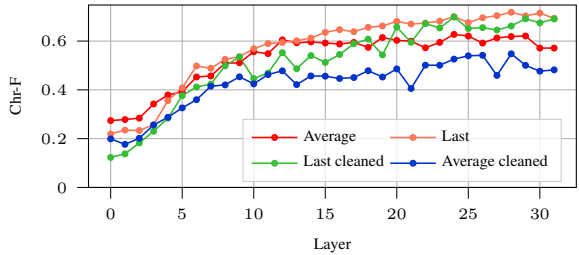
\begin{figure}[hb]
    \centering
    % This file was created with tikzplotlib v0.10.1.
\begin{tikzpicture}

\tikzstyle{every node}=[font=\tiny]
\definecolor{coral25111684}{RGB}{251,116,84}
\definecolor{darkgray176}{RGB}{176,176,176}
\definecolor{lightgray204}{RGB}{204,204,204}
\definecolor{mediumblue}{RGB}{0, 51, 204}
\definecolor{mediumseagreen}{RGB}{51, 190, 51}

\begin{axis}[
legend cell align={left},
legend style={
  fill opacity=0.8,
  draw opacity=1,
  text opacity=1,
  at={(0.97,0.03)},
  anchor=south east,
  draw=lightgray204,
  legend columns=2,
},
width = 8cm,    % Set the desired width
height= 4cm,    % Set the desired heigh
tick align=outside,
tick pos=left,
x grid style={darkgray176},
xlabel={Layer},
xmajorgrids,
xmin=-1.55, xmax=32.55,
xtick style={color=black},
y grid style={darkgray176},
ylabel={Chr-F},
ymajorgrids,
ymin=0, ymax=0.747869976608966,
ytick style={color=black}
]
\addplot [semithick, red, mark=*, mark size=1, mark options={solid}]
table {%
0 0.274029882629908
1 0.278284781590557
2 0.28404598183305
3 0.342092280482206
4 0.379110746945854
5 0.39175771617178
6 0.452875561327231
7 0.457026489677338
8 0.509352154865558
9 0.510663009295447
10 0.556144910709546
11 0.548418919710617
12 0.604966539132148
13 0.592829764285579
14 0.596537939162895
15 0.59208253411481
16 0.587983732611587
17 0.594348652770537
18 0.574187785113658
19 0.614494836803011
20 0.603042168827957
21 0.599570437310119
22 0.572557980698503
23 0.59461685188093
24 0.627355507155266
25 0.620621883283442
26 0.591817029084847
27 0.612963955518798
28 0.618598431557214
29 0.620866597387682
30 0.571207428709319
31 0.571286359395753
};
\addlegendentry{Average}
\addplot [semithick, coral25111684, mark=*, mark size=1, mark options={solid}]
table {%
0 0.219013838723095
1 0.234894166846121
2 0.233380419737328
3 0.257164111165114
4 0.355918505169836
5 0.407862187708815
6 0.497947652159532
7 0.48845564038261
8 0.525190323412619
9 0.535775412911745
10 0.568897095137261
11 0.589602179413717
12 0.594299694351051
13 0.601378939525119
14 0.612456262464303
15 0.635886948925077
16 0.646856020349115
17 0.638612333304161
18 0.656546949335969
19 0.662019978347816
20 0.68036873675979
21 0.669827151828483
22 0.673703013637266
23 0.68173693941987
24 0.700329438055576
25 0.675981852094167
26 0.695773227840139
27 0.704183988044217
28 0.718109890901559
29 0.70337820563441
30 0.714636714631618
31 0.693352762838778
};
\addlegendentry{Last}
\addplot [semithick, mediumseagreen, mark=*, mark size=1, mark options={solid}]
table {%
0 0.122908176753417
1 0.1376745773545
2 0.182377215273535
3 0.230020003437533
4 0.283849415061552
5 0.376016773973035
6 0.411865871107537
7 0.423296002764943
8 0.498472114238732
9 0.534982116138084
10 0.446422348151511
11 0.469004925469799
12 0.552425678491654
13 0.486661395162729
14 0.540517065574013
15 0.512496398133443
16 0.544997138708469
17 0.5892251296948
18 0.607850137495863
19 0.543355299996681
20 0.657457089698619
21 0.594640755393196
22 0.671522540878106
23 0.653781974565321
24 0.698079100038475
25 0.651537684464564
26 0.655244613141817
27 0.645266540171577
28 0.661710025517536
29 0.691585226985089
30 0.674574254099346
31 0.690598085754754
};
\addlegendentry{Last cleaned}
\addplot [semithick, mediumblue, mark=*, mark size=1, mark options={solid}]
table {%
0 0.198976275520151
1 0.176581039146859
2 0.201610142661673
3 0.256002011762582
4 0.28753834881911
5 0.326059563732505
6 0.359720509365208
7 0.414806757579727
8 0.420387456789822
9 0.45354488770465
10 0.424497341236319
11 0.462724268846554
12 0.47806404083957
13 0.42161198941282
14 0.457021908155751
15 0.456018185080854
16 0.446662290772751
17 0.450453426001731
18 0.478551559243678
19 0.452594580080603
20 0.485769784951469
21 0.405662824581145
22 0.501120135180988
23 0.500896116497235
24 0.526032144650164
25 0.539640632714183
26 0.541625431446125
27 0.459169632767656
28 0.547732945194706
29 0.500480301008106
30 0.476496247256956
31 0.481957750121304
};
\addlegendentry{Average cleaned}

\end{axis}

\end{tikzpicture}
    \vspace{-1.2cm}
    \caption{Linear relation decoding on the \texttt{Landmarks\_to\_country} dataset (links landmarks to their home country, \textit{e.g} ``\texttt{Eiffel tower}'' and ``\texttt{France}''). 
    Averaging or cleaning the representations degrades the semantic information of raw representations from the last token of the entity mention.
    %Training on 50 subject and object representations $(\bz_s^\ell,\bz_o^\ell)$ yields up to 72\% Chr-F on test set. 
    }
    \label{fig:RelationDecoding}
    \vspace{-.2cm}
\end{figure}

\begin{figure*}[t]
    \centering
    \vspace{-.2cm}
    \includegraphics[width=1\linewidth]{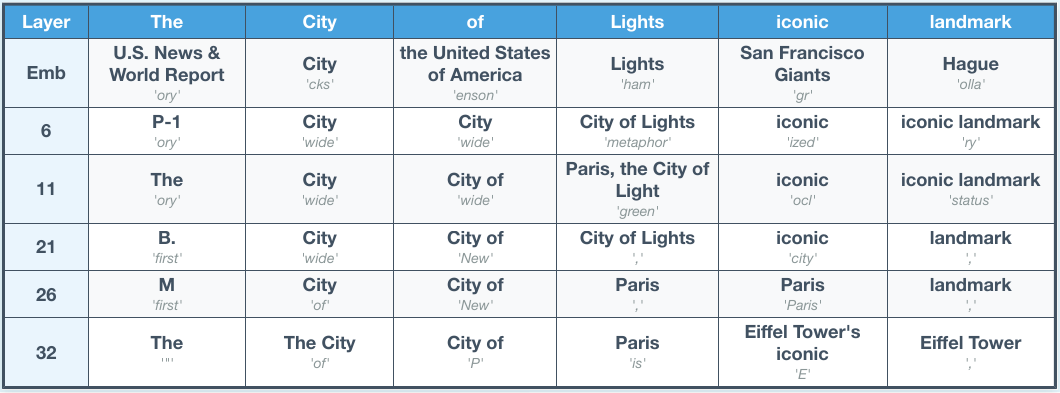}
    \caption{Example application of the \textit{Entity Lens} on the sentence ``\texttt{The City of Lights iconic landmark}'', applied with with the task vectors trained on representations from \textsc{Phi-2} in the \hyperref[par:UncontextualDecoding]{uncontextual} setup. \textsc{Phi-2} associates ``\texttt{City of Lights}'' with Paris, ``\texttt{landmark}'' with the Eiffel Tower in this context. \textit{The token predicted with the logit lens is also shown below.} Additional examples can be found in \Cref{sec:EntityLensViz}.}
    \label{fig:EntityLens_NoContext}
    \vspace{-.2cm}
\end{figure*}

%Logit lens ne fait qu'appliquer la matrice d'unemBedding, pour prédire les token qui aurait été prédits par le modèle d'une part, cela permet uniquement de prédire un seul token. D'autre part, on prédit, on reste sur la tâche de LM là où nous ce qu'on fait, c'est de réellement prédire le la mensonge d'identité.

%%%%%%%%%%%%%% à introduire avant.
% 
We reproduce the idea from \citet{hernandezLinearityRelationDecoding2024} that for some basic relations, the association between $\bz_s$ and $\bz_o$ can  be approximated by a linear model.
Our work extends those results by properly accounting for multi-token entities.
% An idea already explored previously by \citep{Bordes2013TranslatingEF} that modeled relations in knowledge bases as translations between the representations of entities.
We train a linear model $\mathcal L : \bz_s \mapsto W \bz_s + b$ that aims to project the subject representation $\bz_s$ on the object representation $\bz_o$, i.e. $\mathcal L(\bz_s) \approx \bz_o$.

%All details of experiment, will probably be moved to appendix
\paragraph{Experimental settings} 
We use datasets from \citet{hernandezLinearityRelationDecoding2024}
%\footnote{datasets are available at \hyperlink{https://lre.baulab.info/data/}{https://lre.baulab.info/data/}}
for which there are enough samples to train our linear model. Following their methodology, The data is filtered to keep only samples for which the relation is encoded in the model's parametric memory, guaranteeing that the object entity is represented somewhere.
We then optimize our linear model parameters $(W,b)$ using mean-square error 
%$|\mathcal L(\bz_s) - \bz_o|^2$ 
on 50 training samples with stochastic gradient descent. We perform this procedure for each layer $\ell$, using all studied representations ($\bz^\ell$,$\bar \bz^\ell$, $\mathcal{C}_{\bz^\ell}$ or $\mathcal{C}_{\bar \bz^\ell}$) for subject and object entities, allowing to identify which representations carry the most semantic information about an entity.
%as well as assessing the quality of the representation from another point of view.
% they might say, why not a linear regression ?!

\paragraph{Results}
%We follow the same procedure for decoding the entity mention than in \Cref{sec:Background}.
Our mention generation method (See \Cref{sec:Background}) allows the generation of a mention for the obtained representation and then use \textit{exact match} as a metric instead of only comparing the first token only as in \citep{hernandezLinearityRelationDecoding2024, gevaDissectingRecallFactual2023}. In practice, we present the results using the Character F-score (Chr-F), which is not binary and thus produces smoother plots.
\Cref{fig:RelationDecoding} shows that, training a linear model on only 50 samples leads to over 72\% Chr-F (74\% EM) on the \texttt{Landmarks\_to\_country} test set. We show the application to other relation datasets with similar results in appendix (\Cref{fig:OtherRelExtractionXPs}).
Moreover, if using \hyperref[par:average]{average} or \hyperref[par:cleaned]{cleaned} representations yields better performance on mention decoding, this seems to be at the price of losing semantic information, particularly when it comes to encoding relations to other entities -- as shown by the fact that averaging or cleaning representations degrades the results.

\paragraph{Entity Lens} Thanks to our task vectors, we can generate a mention from representations of any token in a text. This allows visualizing which entity the model is ``thinking'' about when processing a token. We name this method the \textit{Entity Lens}, generalizing the \textit{logit lens} \cite{nostalgebraistInterpretingGPTLogit2020}, that associates multi-token mentions with any given representation, using the learned task vectors from \Cref{sec:MentionDecoding}.
For instance, applying the \textit{Entity Lens} to the sentence ``\texttt{The City of Lights iconic landmark}'' shows that the model associates the mention ``\texttt{City of Lights}'' with Paris, and retrieves a representation associated with ``\texttt{Eiffel Tower}'' while processing the token ``\texttt{landmark}''. \Cref{fig:EntityLens_NoContext} shows entities decoded from various hidden representations from \textsc{Phi-2}.
%We also verify that the task vector trained on representations from a specific layer of \textsc{Phi-2} nicely generalizes to representations from other layers.

\section{Conclusion}
\label{sec:conclusion}
%In this paper, we show we can extract layer-agnostic entity (mention) representations from LLMs, decode them using a simple task vector, and transform them for relation decoding, supporting the existence of an entity representation space within LLMs.We have studied two different settings -- namely, the contextual and uncontextual ones, and have shown that

In this study, we have demonstrated that LLMs develop specialized mechanisms for representing and manipulating entities. Our experiments show that they can effectively be prompted using trained \textit{task vectors} to generate complete mentions from entity representations. When given context, they reliably detect and copy mentions of entities, including those outside their parametric memory. Additionally, we showed that these representations can be semantically manipulated to decode basic relations, extending previous work. A direct application of our methodology, the \textit{Entity Lens}, allows for instance to visualize which entity the model is ``thinking'' about at a given layer.

Overall, our results support the existence of an entity representation space within LLMs. This understanding paves the way for further research into how LLMs handle and manipulate knowledge, potentially leading to more explainable and controllable language models.

%Overall, while there are still some topics that need further investigation, our paper is a first step towards demonstrating the existence of an entity representation space within LLMs.

% that are frequent in the LLM training set are sufficient to completely reconstruct an entire entity mention, even for 5-6 tokens ones.
% They can therefore almost be considered as part of the LLM vocabulary.
%For entity mentions that are less common, the model does not store enough information inside these representations, \textit{even if it could}. %(see the supplementary experiment \Cref{sec:TrainingRepresentations}). 
%This is not a limitation as long as, in their regular operating conditions, LLMs can copy the mention from context. %et ça je le déplace

% With \hyperref[sec:MentionDecoding]{our method} in the contextual setting, when the model can access the context containing the entity mention, the model is able to retrieve the right mention with a near-optimal score, providing evidence that the representation can in that case be robustly associated with the named entity mention. 

% LIMITATIONS NOT INCLUDED IN THE 8 PAGES :
% "unlimited extra space after the conclusion for limitations (required, see below) and optionally ethical considerations"  

\section{Limitations}\label{sec:limitations}

\paragraph{Generalization of Task Vectors}

Our method also  assumes that the representations are layer-agnostic, meaning the LLM operates within a single representation space. More importantly, we assume that the task vector -the only learned parameters- is sufficient to instruct the LLM to decode the entity mention without providing any further information.

This discrepancy between contextual and uncontextual settings is reflected by the fact that learned task vectors are not the same across layers (see \Cref{fig:TVsimilarities} in appendix), although we see that they tend to generalize well to other layers (see in appendix, \Cref{fig:EntityLens_NoContext_20}). This design choice was motivated by the known performance of \textit{prompt tuning} \citep{lesterPowerScaleParameterEfficient2021}, which successfully trains embedding vectors to prompt the model into performing specific tasks.

\paragraph{Entity Lens}
Our experiments with task vectors, trained specifically on entities, extend the logit lens by enabling multi-token generation of entity mentions from any representation within the transformer. While our current implementation serves as a preliminary demonstration of this capability, further research is needed to fully explore its potential. 

Currently, our approach generates only a single mention, whereas the traditional logit lens can retrieve the top-$k$ mappings. This limitation could be addressed by employing beam-search generation, which would allow us to generate the "top-$k$" entity mentions for a given representation, thereby enhancing the versatility and applicability of our method. Additionally, the task vectors we train are layer-specific, whereas a more practical implementation would leverage one generalist task vector, which we leave for future work.

%%%%%%%%%%%%%%%%%%%%%%%%%%%%%%%%%%%%%%%%%%%%%%%%%%%%%%%%%%
%%%%%%%%%%%%%%% TO BE INCLUDED AFTER REVIEW  %%%%%%%%%%%%%%%
%%%%%%%%%%%%%%%%%%%%%%%%%%%%%%%%%%%%%%%%%%%%%%%%%%%%%%%%%%
\section{Acknowledgements}
The authors acknowledge the ANR – FRANCE (French National Research Agency) for its financial support of the GUIDANCE project n°ANR-23-IAS1-0003 as well as the Chaire Multi-Modal/LLM ANR Cluster IA ANR-23-IACL-0007. This work was granted access to the HPC resources of IDRIS under the allocation 2024-AD011015440R1 made by GENCI.

\begin{comment}
\section{Ethical Statement}
\end{comment}
%%%%%%%%%%%%%%%%%%%%%%%%%%%%%%%%%%%%%%%%%%%%%%%%%%%%%%%%%%%%

\newpage
\bibliography{Recherche}

\newpage
\appendix

\section{Reproducibility statement}\label{sec:Reproductibility}

All Task vector training have been trained using the CoNLL2003 NER Dataset, as detailed in \Cref{subsec:Dataset}. We performed 15 epochs on the train split using Adam Optimizer. \cite{kingma2017adammethodstochasticoptimization}

We provide a repository where we provide a rendered demo notebook, training code, all hyper-parameters as well as some checkpoints.\footnote{\url{https://github.com/VictorMorand/EntityRepresentations}}  We used the \texttt{transformer lens} \cite{nanda2022transformerlens}, a wrapper around the \texttt{transformers} library \cite{wolf2020huggingfacestransformersstateoftheartnatural}. 

All experiments were conducted on cluster nodes with 80GB NVIDIA A100, 16 or 32GB NVIDIA V100 GPUs. %cite experimaestro for camera ready
%The selected datasets used in our experiments \Cref{sec:Relation_extraction} will also be listed, in addition to the one already mentioned in \Cref{subsec:Dataset}.
% This work was granted access to the HPC resources of IDRIS under the allocation 20XX-[numéro de dossier] made by GENCI 

\begin{table}[h]\label{tab:DataStats}
\centering
\begin{tabular}{@{}ccc@{}} 
\toprule
\multicolumn{1}{r}{\textbf{CoNLL Split}} & \multicolumn{1}{l}{\textbf{Train}} & \multicolumn{1}{l}{\textbf{Test}} \\ \midrule
Number of samples                        & 22449                              & 11120                             \\
Number of unique Entities                & 7820                               & 2521                              \\
Mean text length (in tokens)             & 26.4                               & 26.4                              \\
Mean entity mention length               & 3 tok                              & 3 tok                             \\ \bottomrule
\end{tabular}
\caption{Statistics of our dataset, processed from CoNLL2003 \cite{sang2003introductionconll2003sharedtask}}
\end{table}

\section{Additional experiments}

\subsection{Generating random sequences of fixed token length}\label{subsec:Allbaselines}
\begin{figure*}[ht]
    %\hspace{-.5cm}
    \begin{subfigure}[t]{0.49\textwidth}
        \centering
        % This file was created with tikzplotlib v0.10.1.
\begin{tikzpicture}
\tikzstyle{every node}=[font=\tiny]

\definecolor{firebrick2072831}{RGB}{207,28,31}
\definecolor{coral25111684}{RGB}{251,116,84}
\definecolor{peachpuff252201180}{RGB}{252,201,180}

\definecolor{lightgray204}{RGB}{204,204,204}
\definecolor{darkgray176}{RGB}{176,176,176}
\definecolor{mediumblue}{RGB}{0, 51, 204}
\definecolor{mediumseagreen}{RGB}{51, 190, 51}

\begin{axis}[
legend cell align={left},
legend style={
  fill opacity=0.6,
  draw opacity=1,
  text opacity=1,
  anchor=north east,
  draw=lightgray204,
  legend columns=1
},
width = 8.5cm,    % Set the desired width
height= 6cm,    % Set the desired heigh
tick align=outside,
tick pos=left,
x grid style={lightgray204},
xlabel={layer $\ell$ of \textsc{Phi-2} ($0$ is embedding)},
xmajorgrids,
xmin=-1, xmax=33,
xtick style={color=black},
y grid style={darkgray176},
ylabel={Exact Match},
ytick={0,0.1,...,1},
ymajorgrids,
ymin=-0.02, ymax=1,
ytick style={color=black}
]

\addplot [thick, coral25111684, opacity=1, dashed]
table {%
0   0.335498237149749
1   0.46632028205604
2   0.529782891074411
3   0.578771571720171
4   0.602152532937465
5   0.615327519020226
6   0.617925403599926
7   0.62293560957506
8   0.62293560957506
9   0.621636667285211
10  0.625719057339024
11  0.618482093152719
12  0.625347930970495
13  0.62293560957506
14  0.614214139914641
15  0.618482093152719
16  0.616069771757283
17  0.622193356838003
18  0.621079977732418
19  0.624791241417703
20  0.634440526999443
21  0.637223974763407
22  0.634626090183708
23  0.638337353868992
24  0.629430321024309
25  0.629987010577101
26  0.629244757840045
27  0.623863425496382
28  0.616626461310076
29  0.61328632399332
30  0.604008164780108
31  0.594173316014103
32  0.584709593616626
};
\addlegendentry{Entities}

\addplot [thick, red, opacity=1, solid]
table {%
0  0.9558603491271821
1  0.9387001477104875
2  0.9402764067127345
3  0.9339530332681018
4  0.9400898652021967
5  0.9269965703086722
6  0.9334828101644245
7  0.9252105002476474
8  0.895864106351551
9  0.9105328376703842
10  0.9037419990152634
11  0.8977525314892566
12  0.9156926137776672
13  0.9101959811461175
14  0.8993338267949667
15  0.8924034869240348
16  0.89067922657412
17  0.9032258064516129
18  0.898543569489015
19  0.8933333333333333
20  0.8707214971681851
21  0.8924440068914595
22  0.8848245180425112
23  0.8907211420132907
24  0.8904820766378245
25  0.8621031746031746
26  0.8901098901098901
27  0.8775258748151799
28  0.8697252208047105
29  0.8277583624563155
30  0.8567901234567902
31  0.8510375494071146
32  0.8477448293047596
};
\addlegendentry{1 token}

\addplot [thick, mediumblue, opacity=1, solid]
table {%
0  0.07078534031413612
1  0.322648671446647
2  0.3454699407281964
3  0.381620931716656
4  0.37568594343604894
5  0.3891584053997047
6  0.3745263157894737
7  0.3658485420599958
8  0.36416910841356215
9  0.3513853904282116
10  0.34479848069212915
11  0.3524984319464771
12  0.3448709880427942
13  0.31986531986531985
14  0.3123160992013451
15  0.3238576542429985
16  0.3022670025188917
17  0.3068516183270282
18  0.2952240067624683
19  0.30903728670739417
20  0.28037974683544303
21  0.28520084566596193
22  0.2921940928270042
23  0.24506094997898276
24  0.2809060118543607
25  0.28716645489199494
26  0.22111511553953783
27  0.251892346509672
28  0.244794952681388
29  0.2297154899894626
30  0.23226351351351351
31  0.23653886444584118
32  0.2120133947258267
};
\addlegendentry{2 tokens}

\addplot [thick, mediumseagreen, opacity=1, solid]
table {%
0  0.007785289899610736
1  0.02771504824471361
2  0.06994290375203915
3  0.09109477124183006
4  0.10052910052910052
5  0.09752099979512395
6  0.10772066352652059
7  0.09681372549019608
8  0.09555964804583589
9  0.08627370997348562
10  0.08704536166734778
11  0.09405537459283388
12  0.09317531671434409
13  0.08081632653061224
14  0.0895675343308055
15  0.07978831671076735
16  0.07879159012043274
17  0.07554469558134799
18  0.07830709466366796
19  0.07113738279657562
20  0.06442405708460755
21  0.07430087773014901
22  0.07538304392236976
23  0.06902218570254724
24  0.07430213464696224
25  0.06948145111703218
26  0.06766612641815235
27  0.07225964482547459
28  0.05728848114169215
29  0.0678069639584606
30  0.06744091279543603
31  0.05878750765462339
32  0.05027590435315757
};
\addlegendentry{3 tokens}

\end{axis}

\end{tikzpicture}
        \vspace{-1cm}
        \subcaption{\hyperref[par:UncontextualDecoding]{Uncontextual mention decoding} results by layer.}
        \label{figsub:AllBaselinesNoContext}
    \end{subfigure}
    %\hspace{-.5cm}
    \hfill
    \begin{subfigure}[t]{0.49\textwidth}
        \centering
        % This file was created with tikzplotlib v0.10.1.
\begin{tikzpicture}
\tikzstyle{every node}=[font=\tiny]

\definecolor{firebrick2072831}{RGB}{207,28,31}
\definecolor{coral25111684}{RGB}{251,116,84}
\definecolor{peachpuff252201180}{RGB}{252,201,180}

\definecolor{lightgray204}{RGB}{204,204,204}
\definecolor{darkgray176}{RGB}{176,176,176}
\definecolor{mediumblue}{RGB}{0, 51, 204}
\definecolor{mediumseagreen}{RGB}{51, 190, 51}

\begin{axis}[
legend cell align={left},
legend to name=empty, % This removes the legend
width = 8.5cm,    % Set the desired width
height= 6cm,    % Set the desired heigh
tick align=outside,
tick pos=left,
x grid style={darkgray176},
xlabel={layer $\ell$ of \textsc{Phi-2} ($0$ is embedding)},
xmajorgrids,
xmin=-1, xmax=33,
xtick style={color=black},
y grid style={lightgray204},
ytick={0,0.1,...,1},
ymajorgrids,
ymin=-0.02, ymax=1,
ytick style={color=black}
]

\addplot [thick, coral25111684, opacity=1, dashed]
table {%
0 0.798014473928373
1 0.883806519453207
2 0.902331910682254
3 0.908053442197068
4 0.910836889961032
5 0.910001855631843
6 0.91439351765943
7 0.919094451660791
8 0.915352260778128
9 0.917176965423393
10  0.921444918661471
11  0.926826251005134
12  0.925001546359869
13  0.924970619162492
14  0.92475412878085
15  0.924939691965114
16  0.927259231768417
17  0.928867446032041
18  0.931310694624853
19  0.928589101255644
20  0.931310694624853
21  0.931063277045834
22  0.928805591637286
23  0.928527246860889
24  0.925403599925775
25  0.924537638399208
26  0.92416651203068
27  0.920548029937527
28  0.914115172883033
29  0.906197810354426
30  0.902331910682254
31  0.894785674522175
32  0.889095070204738
};
\addlegendentry{Entities}

\addplot [thick, red, opacity=1, solid]
table {%
0  0.9541353383458646
1  0.9489467162329616
2  0.931377424167081
3  0.9219367588932806
4  0.9038748137108793
5  0.8927429274292743
6  0.887598814229249
7  0.8448361469712016
8  0.8332099679249938
9  0.8925097276264592
10  0.8955075701166543
11  0.8848335388409371
12  0.8787281242297263
13  0.8893606393606394
14  0.8429463171036204
15  0.8383712215838122
16  0.8934385791810557
17  0.8002472187886279
18  0.8842079329884208
19  0.8554784333168072
20  0.8406971035837015
21  0.8585161449346808
22  0.876509736258319
23  0.8850574712643678
24  0.8426633785450062
25  0.8704066634002939
26  0.8556573409259718
27  0.8732186732186732
28  0.8464206093633887
29  0.8667997006734847
30  0.8528971028971029
31  0.8441780821917808
32  0.8504580341668729
};
\addlegendentry{1 token}

\addplot [thick, mediumblue, opacity=1, solid]
table {%
0  0.24071729957805907
1  0.29508542501581947
2  0.3676254461473861
3  0.37175188600167647
4  0.42589473684210527
5  0.39701241321270775
6  0.39196297075531245
7  0.38114321873022566
8  0.408
9  0.39936908517350156
10  0.391578947368421
11  0.35
12  0.35932851678708033
13  0.3608704838368899
14  0.37801552338997274
15  0.3577133249264397
16  0.3334036271615352
17  0.36216670165861853
18  0.3374580536912752
19  0.3303834808259587
20  0.34403188588210615
21  0.34203821656050953
22  0.3443445536274305
23  0.3447623054270088
24  0.33591298891445304
25  0.3264830508474576
26  0.29710298160287585
27  0.3068325601012231
28  0.30403700588730026
29  0.2865558594571849
30  0.2771819137749737
31  0.23128679562657695
32  0.23819517313746066
};
\addlegendentry{2 tokens}

\addplot [thick, mediumseagreen, opacity=1, solid]
table {%
0  0.12097103223174215
1  0.15507364975450083
2  0.17583316295236148
3  0.18670756646216768
4  0.1738599348534202
5  0.18387293830177154
6  0.18818737270875763
7  0.18612244897959185
8  0.19771708112515288
9  0.19983719983719983
10  0.19657771440211855
11  0.1939840392879067
12  0.20427263479145474
13  0.18966218966218967
14  0.18803418803418803
15  0.1829543126408523
16  0.18128298822574096
17  0.19726083401471792
18  0.18207739307535642
19  0.1825010150223305
20  0.1920408163265306
21  0.18143031288094272
22  0.1805527123848516
23  0.17780040733197555
24  0.16327788046826863
25  0.17903752039151713
26  0.16812895327484187
27  0.16331505179768435
28  0.17822990844354017
29  0.17038705713700594
30  0.14805725971370143
31  0.11745513866231648
32  0.124185667752443
};
\addlegendentry{3 tokens}

\end{axis}

\end{tikzpicture}
        \vspace{-1cm}
        \subcaption{\hyperref[par:contextualDecoding]{Contextual mention decoding} results by layer.}
        \label{figsub:AllBaselinesContext}
    \end{subfigure}
    \caption{Baseline -- Decoding random sequences of fixed token length compared to decoding entity mentions with \textsc{Phi-2} (end token is constrained to be the end of a word). If the model can easily retrieve one represented token, manipulating entity mentions is a more natural task then manipulating tokens. The observed behavior is similar on the other models considered.}
    \label{fig:AllBaselines}
    %\vspace{-0.5cm}
\end{figure*}
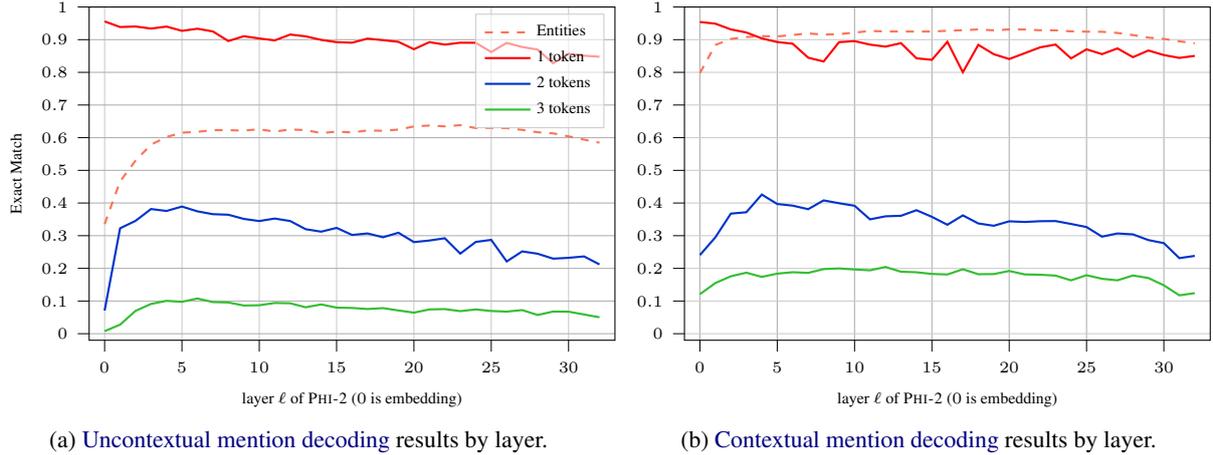

We provide \Cref{fig:AllBaselines} all the results obtained for our control experiment described in \Cref{par:baseline}

\subsection{Generalization of learned Task Vectors}\label{subsec:TV_generalization}

\begin{figure}
    \begin{subfigure}[b]{0.45\textwidth}
        \centering
        \includegraphics[width=\textwidth]{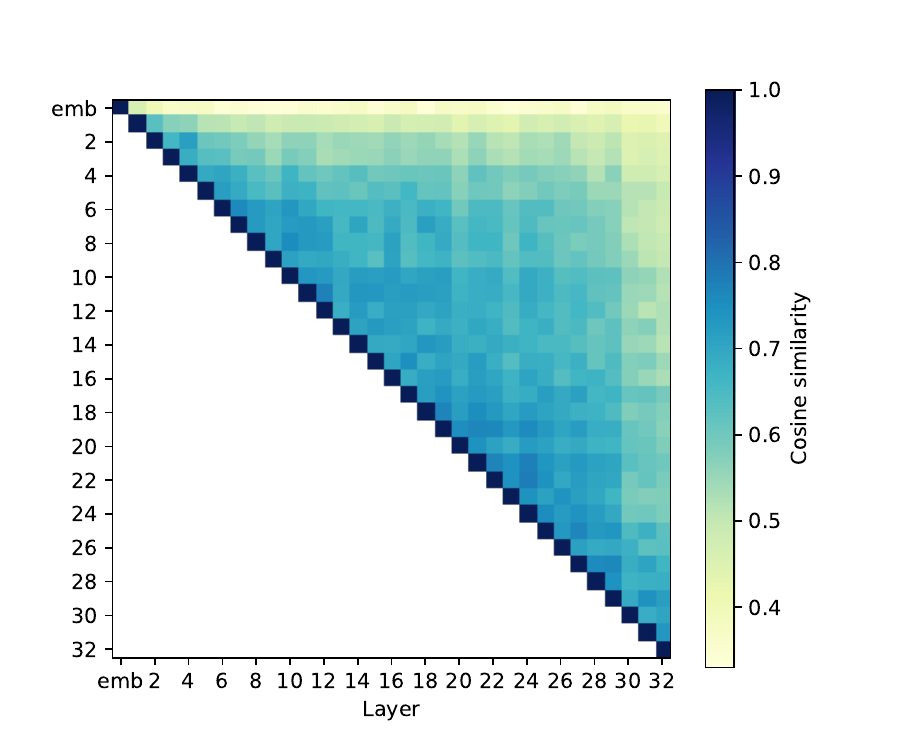}
        \subcaption{Similarities between trained task vectors in the \hyperref[par:UncontextualDecoding]{uncontextual} setup}
    \end{subfigure}
    \begin{subfigure}[b]{0.45\textwidth}
        \centering
        \includegraphics[width=\textwidth]{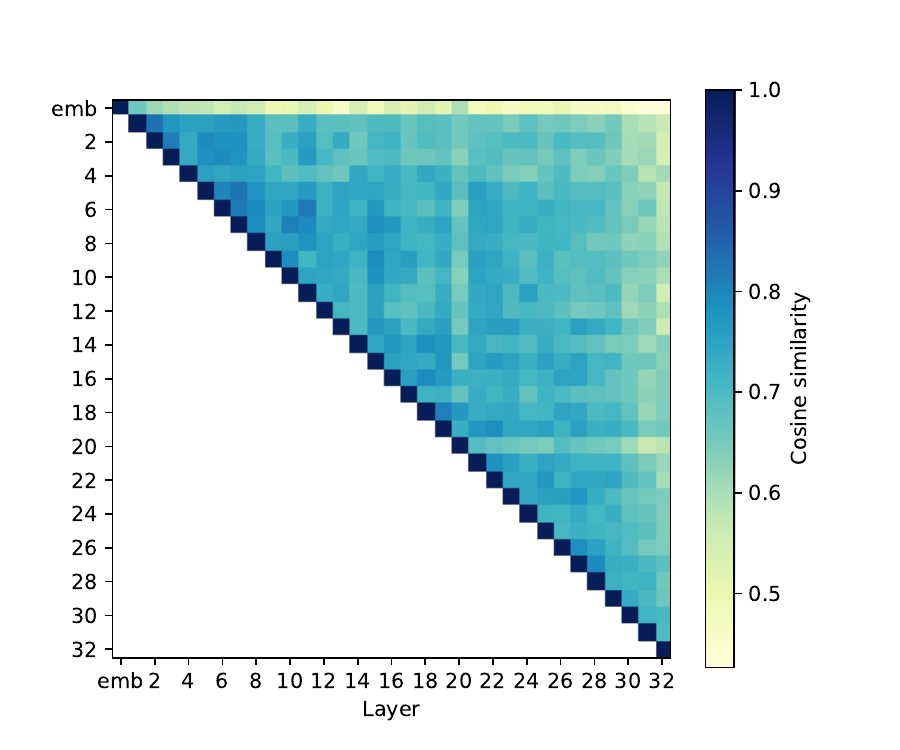}
        \subcaption{Similarities between trained task vectors in the \hyperref[par:contextualDecoding]{contextual} setup}
        %\vspace{-.6cm}
    \end{subfigure}
    %\vspace{-1cm}
    \caption{Cosine Similarity comparison of all trained task Vectors for \textsc{Phi-2}. Training at each layer seems to lead to a different task vector, although they are shown to generalize well (see \Cref{subsec:TV_generalization}).}
    \label{fig:TVsimilarities}
\end{figure}

\paragraph{Other layers} Each task vectors has been trained with representations from a specific transformer layer, this methodology allows to further analyze the impact of the layer on the quality of the representations. Even if they are not particularly similar (cf \Cref{fig:TVsimilarities}), task vectors generalize well to other layers. This can be seen in \Cref{fig:EntityLens_NoContext_20}, where we apply the \textit{Entity Lens} using the same task vector $\theta_{20}$ for all layers. The generated mentions are still consistent even if the representations are not extracted at layer $20$.

\paragraph{Different setups} Despite being similar, the two setups may imply a different generation mechanism from the model. In the \hyperref[par:UncontextualDecoding]{uncontextual setup}, we require the model to retrieve a mention from its parametric memory, whereas in the \hyperref[par:contextualDecoding]{contextual setup} we only require it to copy the right mention. 
Evaluating task vectors across tasks (\Cref{tab:TVGeneralization}) shows that, despite a ~50\% drop in performance, task vectors trained for one setting can be used in the other.
This demonstrates that while the two processes are different, they however have some similarities. Investigating which is left for future work.

%Evaluation of task vectors trained for one task on the other setting (\Cref{tab:TVGeneralization}) show that (with around 50\% drop in performance) task vectors trained for uncontextual decoding can be used for contextual decoding (and vice-versa). 
% some generalization capabilities (\textit{e.g} the contextual task vector $\theta_{20}$ still yields 41\% exact matching when evaluated in an uncontextual setting, see \Cref{tab:TVGeneralization}). 

\begin{table}[h]
\centering
\resizebox{\columnwidth}{!}{%
\begin{tabular}{ccccc}
\hline
\multirow{2}{*}{Task Vector} & \multicolumn{2}{c}{Uncontextual evaluation} & \multicolumn{2}{c}{Contextual evaluation} \\
                            & Exact match & Chr-F & Exact match & Chr-F \\ \hline
Phi-2 $\ell \ 20$, uncontextual & 64\%        & 61\%  & 15\%        & 30\%  \\
Phi-2 $\ell \ 20$, contextual   & 41\%        & 40\%  & 93\%        & 94\%  \\
Random Vector               & 0\%         & 13\%  & 0\%         & 17\%  \\ \hline
\end{tabular}%
}
\caption{Despite being trained for distinct tasks, task vectors do show some generalization to other generation settings.}\label{tab:TVGeneralization}
\vspace{-.5cm}
\end{table}

\subsection{Analyzing entity representations}\label{sec:reprConstruction}

%After studying to what extent a representation can be bound to a named entity, we 
Here, we explore how entity mention representations are built inside an LLM.

\begin{figure}[htb]
    \centering
        \includegraphics[width=0.5\textwidth]{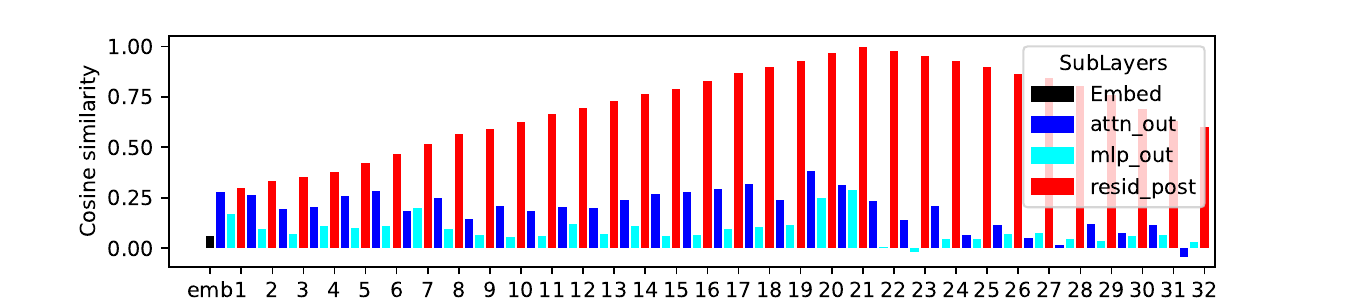}
        \vspace{-.5cm}
        \caption{Similarity between the last token representations ($\bz^\ell$) and intermediate representations from \textsc{Phi-2}. Different sublayers are shown, including outputs from the MLP and Multi Head Self-Attention (Attn). Application to a specific prompt : \texttt{'Port|ugal| called| up| Port|o| central| defender| Jo|ao| Manuel| P|\textbf{into}'}. The observed behavior is representative of the general one.}
        \vspace{-.5cm}
    \label{fig:RepresentationConvergence}
\end{figure}

\paragraph{Successive representations} We  explore the convergence of the successive representations to the one that we extract at layer $\ell$. We consider the cosine similarity of the entity representation $\bz^\ell$ with the output from different layers, including Multi-head self-attention (Attn) and Multi-Layer Perceptron (MLP) sublayers. %-> without mentionning tokens:
 Sublayers iteratively \textit{add} or \textit{remove} information in the \textit{residual stream}. %The results show t
 There is no predominant layer for the construction of $\bz^\ell$ (See \Cref{fig:RepresentationConvergence}).
 %We cannot however spot in \Cref{fig:RepresentationConvergence}

%For a multi-token entity mention $e = (t_{e_1},...,t_{e_2})$ in a given context $(t_1,...,t_n)$, we dissect the construction of the representation $\bz^\ell_n, \ell \in \{1, ..., N_L\}$ across the layers of the transformer model. We compute the similarity between the intermediate representation of token $t_k$ at layer $\ell$: $\bz_k^\ell$ and the extracted representation of the last token $\bz_{e_2}^\ell$.
    
\begin{comment} simpler is better, proj on vocab actually don't change the ccl.
$$\mathcal{S}:j \mapsto \cos(\bz^j,\bz^\ell)$$
\paragraph{Remark: Limitations of cosine similarity usage}
If using cosine similarity is widely used to compare vectors, it is limited in the scope of transformer models. The distribution of representation vectors in a transformer model \textit{is not isotropic}. According to \citet{ethayarajhHowContextualAre2019}, it resembles a cone.

To overcome this limitation, we therefore compute the similarity \textit{in vocabulary space} :
$$\mathcal{S^V}:k,l \mapsto \cos(W_U \bz_k^\ell, W_U \bz_n^\ell)$$
\end{comment}

\paragraph{Causal analysis}
is an alternative mean to assess the contribution of one component of a transformer layer on the representation construction.
We use \textit{sublayer knockout}, as done in \cite{gevaDissectingRecallFactual2023}, by zeroing out the output of one MLP or attention block of a given layer $\ell$ while computing the representation $\bz^\ell$. We can measure how much this block contributes to this representation by comparing the similarity of the obtained representation with the original one. We observe that, apart from the first layer, no other blocks have a causal effect on the final representation $\bz^\ell$, confirming the observation made above.
% In practice, we compare the similarity of the obtained representation when knocking out Attn or mlp in different layers with the original representation (without knockout).

%\begin{figure}[ht]
%    \centering
%    \includegraphics[scale=.5]{images/RawReprConvergence.png}
%    \caption{Raw Cosine similarities between successive representations and extracted one}
%    \label{fig:RawReprSims}
%\end{figure}
% \subsubsection{Construction without context}
% Difference of representation with context / without context.
% \\todo[inline]{We could do as in \\cite{ethayarajhHowContextualAre2019}, compute representations with and without context, and then compute : }
% \\begin{itemize}
%     \\item 
%     The average similarity btw token representation in context / without
%     \\item
%     The MEV ? we'd need many different contexts for each
% \\end{itemize}

 \paragraph{Conclusion}
Both our similarity and causal analysis experiments lead us to conclude that, as previous work also suggested \cite{mengLocatingEditingFactual2022, gevaDissectingRecallFactual2023}, there is no clear location where the representation of an entity is ``completed''. The construction of entity representations are a \textit{a smooth, iterative, and massively superposed process.} This aligns with recent contributions on superposition and feature disentanglement, and may be explained by the use of dropout during LLM pretraining, nudging the model to develop redundant circuits~\cite{elhageToyModelsSuperposition2022, brickenMonosemanticityDecomposingLanguage2023}.

\subsection{Training Representations}\label{sec:TrainingRepresentations}
We explore in this section what features in the representation are really used to generate a mention. To obtain the minimum information required to retrieve the mention, we optimize blank noise to make the model retrieve the right mention when prompted with a task vector trained in the context of our \hyperref[par:UncontextualDecoding]{uncontextual mention generation} experiment. By doing it several times and averaging the obtained vectors, we hope to keep only hat is needed to regenerate the mention.
\paragraph{Conclusion} Our findings demonstrate that it is possible to train a vector to encode almost any mentions, within reasonable token limits. This confirms that the transformer's latent space has the capacity to store numerous tokens. LLMs however typically do not utilize this capability when they can access the context, as they can simply copy and paste the appropriate tokens from it.

\section{Textual Examples} 
\begin{table}[ht!]
\centering
\begin{tabular}{@{}cc@{}}
\toprule
\textbf{Original}      & \textbf{Inferred}  \\ \midrule
Roberto Mancini        & Carlo Mazzone      \\
Pierre Van Hooydonk    & Marc D'Haese       \\
Guenther Huber         & Peter Huber        \\
Wenchang               & Changsha           \\
Michael Cornwell       & Mark Calwell       \\
IGLS                   & GLIS               \\
Ole Einar Bjorndalen   & Bjorn Dæhlie       \\
Alba Berlin            & Berlin             \\
John Langmore          & John Molyneaux     \\
Patasse                & Passeau            \\
Rangoon                & Yangon             \\
M. Waugh               & J. Waugh           \\
Major                  & John Major         \\
Lahd                   & Hlad               \\
WARSAW                 & WAWRZAWA           \\
Kim Pan Keun           & Lee Dong-kook      \\
David Boon             & J. Boon            \\
Berisha                & Bushi              \\
Gunn Margit Andreassen & Ingrid Bjørnson    \\
Abel Balbo             & Giuseppe Bologna   \\ \bottomrule
\end{tabular}%
\caption{Examples of failed generations sampled randomly from the the uncontextual mention generation results (\Cref{figsub:UncontextResults}) for \textsc{Phi-2} at layer 15.}
\label{tab:uncontextualized_samples}
\end{table}

\begin{table*}[ht]
\resizebox{\textwidth}{!}{%
\begin{tabular}{@{}cccccccccc@{}}
\toprule
            & \textbf{n\_params} & \textbf{n\_layers} & \textbf{d\_model} & \textbf{n\_heads} & \textbf{act\_fn} & \textbf{n\_ctx} & \textbf{d\_vocab} & \textbf{d\_head} & \textbf{d\_mlp} \\ \midrule
\textsc{phi-1.5    } & 1.2B               & 24                 & 2048              & 32                & gelu             & 2048            & 51200             & 64               & 8192            \\
\textsc{phi-2      } & 2.5B               & 32                 & 2560              & 32                & gelu             & 2048            & 51200             & 80               & 10240           \\
\textsc{phi-3      } & 3.6B               & 32                 & 3072              & 32                & silu             & 4096            & 32064             & 96               & 8192            \\ \midrule
\textsc{pythia-160m} & 85M                & 12                 & 768               & 12                & gelu             & 2048            & 50304             & 64               & 3072            \\
\textsc{pythia-410m} & 302M               & 24                 & 1024              & 16                & gelu             & 2048            & 50304             & 64               & 4096            \\
\textsc{pythia-1b  } & 805M               & 16                 & 2048              & 8                 & gelu             & 2048            & 50304             & 256              & 8192            \\
\textsc{pythia-1.4b} & 1.2B               & 24                 & 2048              & 16                & gelu             & 2048            & 50304             & 128              & 8192            \\
\textsc{pythia-2.8b} & 2.5B               & 32                 & 2560              & 32                & gelu             & 2048            & 50304             & 80               & 10240           \\
\textsc{pythia-6.9b} & 6.4B               & 32                 & 4096              & 32                & gelu             & 2048            & 50432             & 128              & 16384           \\ \bottomrule
\end{tabular}%
}%
\caption{Characteristics of the models considered in this work.}
\label{table:modelParams}
\end{table*}

\section{Complementary Figures}

\begin{figure}
    \centering
    \begin{subfigure}[b]{0.45\textwidth}
        % This file was created with tikzplotlib v0.10.1.

% This file was created with tikzplotlib v0.10.1.
\begin{tikzpicture}
\tikzstyle{every node}=[font=\small]

\definecolor{coral25111684}{RGB}{251,116,84}
\definecolor{darkgray176}{RGB}{176,176,176}
\definecolor{darkslateblue3654149}{RGB}{36,54,149}
\definecolor{firebrick2072831}{RGB}{207,28,31}
\definecolor{lightgreen151214184}{RGB}{151,214,184}
\definecolor{lightseagreen36152192}{RGB}{36,152,192}
\definecolor{mediumturquoise81188193}{RGB}{81,188,193}
\definecolor{palegoldenrod214239178}{RGB}{214,239,178}
\definecolor{peachpuff252201180}{RGB}{252,201,180}
\definecolor{steelblue33100171}{RGB}{33,100,171}

\begin{axis}[
tick align=outside,
tick pos=left,
title={ \textbf{Uncontextual} mention Decoding  on CoNLL2003},
x grid style={darkgray176},
xlabel={Model size (B parameters)},
width = 7cm,    % Set the desired width
height= 5cm,    % Set the desired heigh
xmajorgrids,
xmin=-0.152, xmax=8,
xtick style={color=black},
y grid style={darkgray176},
ylabel={Exact Match},
ymajorgrids,
ymin=0.310270922249026, ymax=0.685646687697161,
ytick style={color=black}
]
\addplot [semithick, firebrick2072831, opacity=0.5, dashed]
table {%
1.5 0.511226572647987
2.6 0.64872889218779
3.8 0.664872889218779
};
\addplot [semithick, darkslateblue3654149, opacity=0.5, dashed]
table {%
0.16 0.327333457042123
0.302 0.386156986453888
0.805 0.5433290035257
1.2 0.558916311003897
2.5 0.609575060308035
6.4 0.668584152904064
};
\addplot [semithick, peachpuff252201180, mark=*,mark size=1,, mark options={solid}, only marks]
table {%
1.5 0.511226572647987
};
\addplot [semithick, coral25111684, mark=*,mark size=1,, mark options={solid}, only marks]
table {%
2.6 0.64872889218779
};
\addplot [semithick, firebrick2072831, mark=*,mark size=1,, mark options={solid}, only marks]
table {%
3.8 0.664872889218779
};
\addplot [semithick, palegoldenrod214239178, mark=*,mark size=1,, mark options={solid}, only marks]
table {%
0.16 0.327333457042123
};
\addplot [semithick, lightgreen151214184, mark=*,mark size=1,, mark options={solid}, only marks]
table {%
0.302 0.386156986453888
};
\addplot [semithick, mediumturquoise81188193, mark=*,mark size=1,, mark options={solid}, only marks]
table {%
0.805 0.5433290035257
};
\addplot [semithick, lightseagreen36152192, mark=*,mark size=1,, mark options={solid}, only marks]
table {%
1.2 0.558916311003897
};
\addplot [semithick, steelblue33100171, mark=*,mark size=1,, mark options={solid}, only marks]
table {%
2.5 0.609575060308035
};
\addplot [semithick, darkslateblue3654149, mark=*,mark size=1,, mark options={solid}, only marks]
table {%
6.4 0.668584152904064
};
\draw (axis cs:1.53,0.511226572647987) node[
  scale=0.6,
  anchor=base west,
  text=black,
  rotate=0.0
]{phi-1.5};
\draw (axis cs:2.63,0.64872889218779) node[
  scale=0.6,
  anchor=base west,
  text=black,
  rotate=0.0
]{phi-2};
\draw (axis cs:3.83,0.664872889218779) node[
  scale=0.6,
  anchor=base west,
  text=black,
  rotate=0.0
]{phi-3};
\draw (axis cs:0.19,0.327333457042123) node[
  scale=0.6,
  anchor=base west,
  text=black,
  rotate=0.0
]{pythia-160m};
\draw (axis cs:0.332,0.386156986453888) node[
  scale=0.6,
  anchor=base west,
  text=black,
  rotate=0.0
]{pythia-410m};
\draw (axis cs:0.835,0.5433290035257) node[
  scale=0.6,
  anchor=base west,
  text=black,
  rotate=0.0
]{pythia-1b};
\draw (axis cs:1.23,0.558916311003897) node[
  scale=0.6,
  anchor=base west,
  text=black,
  rotate=0.0
]{pythia-1.4b};
\draw (axis cs:2.53,0.609575060308035) node[
  scale=0.6,
  anchor=base west,
  text=black,
  rotate=0.0
]{pythia-2.8b};
\draw (axis cs:6.43,0.668584152904064) node[
  scale=0.6,
  anchor=base west,
  text=black,
  rotate=0.0
]{pythia-6.9b};
\end{axis}

\end{tikzpicture}
        \vspace{-1cm}
    \end{subfigure}
    \begin{subfigure}[b]{0.45\textwidth}
        % This file was created with tikzplotlib v0.10.1.
% This file was created with tikzplotlib v0.10.1.
\begin{tikzpicture}
\tikzstyle{every node}=[font=\small]

\definecolor{coral25111684}{RGB}{251,116,84}
\definecolor{darkgray176}{RGB}{176,176,176}
\definecolor{darkslateblue3654149}{RGB}{36,54,149}
\definecolor{firebrick2072831}{RGB}{207,28,31}
\definecolor{lightgreen151214184}{RGB}{151,214,184}
\definecolor{lightseagreen36152192}{RGB}{36,152,192}
\definecolor{mediumturquoise81188193}{RGB}{81,188,193}
\definecolor{palegoldenrod214239178}{RGB}{214,239,178}
\definecolor{peachpuff252201180}{RGB}{252,201,180}
\definecolor{steelblue33100171}{RGB}{33,100,171}

\begin{axis}[
tick align=outside,
tick pos=left,
title={ \textbf{Contextual} mention Decoding  on CoNLL2003},
x grid style={darkgray176},
xlabel={Model size (B parameters)},
width = 7cm,    % Set the desired width
height= 5cm,    % Set the desired heigh
xmajorgrids,
xmin=-0.152, xmax=8,
xtick style={color=black},
y grid style={darkgray176},
ylabel={Exact Match},
ymajorgrids,
ymin=0.575208758582297, ymax=0.951605121543886,
ytick style={color=black}
]
\addplot [semithick, firebrick2072831, opacity=0.5, dashed]
table {%
1.5 0.839487845611431
2.6 0.934496195954723
3.8 0.796808313230655
};
\addplot [semithick, darkslateblue3654149, opacity=0.5, dashed]
table {%
0.16 0.59231768417146
0.302 0.738912599740212
0.805 0.724438671367601
1.2 0.812581183893116
2.5 0.828168491371312
6.4 0.859343106327705
};
\addplot [semithick, peachpuff252201180, mark=*,mark size=1,, mark options={solid}, only marks]
table {%
1.5 0.839487845611431
};
\addplot [semithick, coral25111684, mark=*,mark size=1,, mark options={solid}, only marks]
table {%
2.6 0.934496195954723
};
\addplot [semithick, firebrick2072831, mark=*,mark size=1,, mark options={solid}, only marks]
table {%
3.8 0.796808313230655
};
\addplot [semithick, palegoldenrod214239178, mark=*,mark size=1,, mark options={solid}, only marks]
table {%
0.16 0.59231768417146
};
\addplot [semithick, lightgreen151214184, mark=*,mark size=1,, mark options={solid}, only marks]
table {%
0.302 0.738912599740212
};
\addplot [semithick, mediumturquoise81188193, mark=*,mark size=1,, mark options={solid}, only marks]
table {%
0.805 0.724438671367601
};
\addplot [semithick, lightseagreen36152192, mark=*,mark size=1,, mark options={solid}, only marks]
table {%
1.2 0.812581183893116
};
\addplot [semithick, steelblue33100171, mark=*,mark size=1,, mark options={solid}, only marks]
table {%
2.5 0.828168491371312
};
\addplot [semithick, darkslateblue3654149, mark=*,mark size=1,, mark options={solid}, only marks]
table {%
6.4 0.859343106327705
};
\draw (axis cs:1.53,0.839487845611431) node[
  scale=0.6,
  anchor=base west,
  text=black,
  rotate=0.0
]{phi-1.5};
\draw (axis cs:2.63,0.934496195954723) node[
  scale=0.6,
  anchor=base west,
  text=black,
  rotate=0.0
]{phi-2};
\draw (axis cs:3.83,0.796808313230655) node[
  scale=0.6,
  anchor=base west,
  text=black,
  rotate=0.0
]{phi-3};
\draw (axis cs:0.19,0.59231768417146) node[
  scale=0.6,
  anchor=base west,
  text=black,
  rotate=0.0
]{pythia-160m};
\draw (axis cs:0.332,0.738912599740212) node[
  scale=0.6,
  anchor=base west,
  text=black,
  rotate=0.0
]{pythia-410m};
\draw (axis cs:0.835,0.724438671367601) node[
  scale=0.6,
  anchor=base west,
  text=black,
  rotate=0.0
]{pythia-1b};
\draw (axis cs:1.23,0.812581183893116) node[
  scale=0.6,
  anchor=base west,
  text=black,
  rotate=0.0
]{pythia-1.4b};
\draw (axis cs:2.53,0.828168491371312) node[
  scale=0.6,
  anchor=base west,
  text=black,
  rotate=0.0
]{pythia-2.8b};
\draw (axis cs:6.43,0.859343106327705) node[
  scale=0.6,
  anchor=base west,
  text=black,
  rotate=0.0
]{pythia-6.9b};
\end{axis}

\end{tikzpicture}%
        \vspace{-1cm}
    \end{subfigure}
    \caption{Aggregated Results comparing best performances depending on model size. Larger models demonstrate greater capability. Performance drop of the \textsc{Phi-3} model can be explained by the use of a significantly smaller vocabulary size (32k vs 50k for all other models considered here) as well as the instruction tuning.}
    \label{fig:AggResultsSizes}
\end{figure}
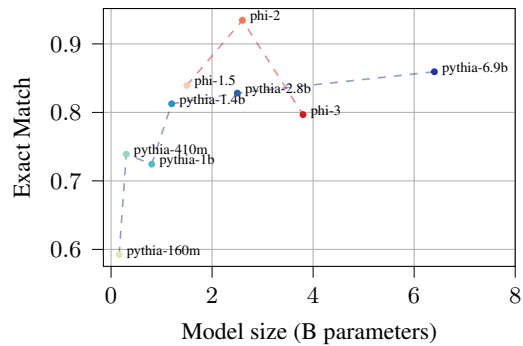

\begin{figure}
    \begin{subfigure}[b]{0.45\textwidth}
        \centering
        % This file was created with tikzplotlib v0.10.1.
\begin{tikzpicture}[scale=0.8]

\tikzstyle{every node}=[font=\small]

\definecolor{coral25111684}{RGB}{251,116,84}
\definecolor{darkgray176}{RGB}{176,176,176}
\definecolor{darkslateblue3654149}{RGB}{36,54,149}
\definecolor{firebrick2072831}{RGB}{207,28,31}
\definecolor{lightgray204}{RGB}{204,204,204}
\definecolor{lightgreen151214184}{RGB}{151,214,184}
\definecolor{lightseagreen36152192}{RGB}{36,152,192}
\definecolor{mediumturquoise81188193}{RGB}{81,188,193}
\definecolor{palegoldenrod214239178}{RGB}{214,239,178}
\definecolor{peachpuff252201180}{RGB}{252,201,180}
\definecolor{steelblue33100171}{RGB}{33,100,171}

\begin{axis}[
legend cell align={left},
legend style={
  fill opacity=0.8,
  draw opacity=1,
  text opacity=1,
  at={(0.03,0.97)},
  anchor=north west,
  draw=lightgray204
},
tick align=outside,
tick pos=left,
title={Exact Match per quantile for Pythias models},
x grid style={darkgray176},
xlabel={Quantile},
xmajorgrids,
xmin=-0.05, xmax=1.05,
xtick style={color=black},
y grid style={darkgray176},
ylabel={Exact Match (\%)},
ymajorgrids,
ymin=-3.47931167826759, ymax=73.0655452436195,
ytick style={color=black}
]
\addplot [semithick, palegoldenrod214239178, mark=*, mark size=1, mark options={solid}]
table {%
0 0
0.1 0
0.2 0
0.3 0.254777070063694
0.4 0.175901495162709
0.5 0.262295081967213
0.6 0.852557673019057
0.7 1.83953033268102
0.8 4.17070805043647
0.9 12.88
1 34.1067285382831
};
\addlegendentry{pythia-160m  l6}
\addplot [semithick, lightgreen151214184, mark=*, mark size=1, mark options={solid}]
table {%
0 0
0.1 0
0.2 0
0.3 0
0.4 0.0879507475813544
0.5 0.0655737704918033
0.6 1.60481444332999
0.7 4.18786692759295
0.8 7.66246362754607
0.9 16
1 35.614849187935
};
\addlegendentry{pythia-410m  l10}
\addplot [semithick, mediumturquoise81188193, mark=*, mark size=1, mark options={solid}]
table {%
0 0
0.1 0
0.2 0
0.3 1.01910828025478
0.4 2.63852242744063
0.5 4
0.6 11.3841524573721
0.7 20.1956947162427
0.8 30.1002263174911
0.9 40.4533333333333
1 56.5931941221964
};
\addlegendentry{pythia-1b  l6}
\addplot [semithick, lightseagreen36152192, mark=*, mark size=1, mark options={solid}]
table {%
0 0
0.1 0
0.2 0.23696682464455
0.3 2.54777070063694
0.4 3.69393139841689
0.5 5.37704918032787
0.6 13.3400200601805
0.7 23.8356164383562
0.8 33.3333333333333
0.9 43.0666666666667
1 58.2173240525909
};
\addlegendentry{pythia-1.4b  l7}
\addplot [semithick, steelblue33100171, mark=*, mark size=1, mark options={solid}]
table {%
0 0
0.1 0
0.2 1.4218009478673
0.3 1.91082802547771
0.4 3.43007915567282
0.5 6.49180327868853
0.6 14.6940822467402
0.7 25.7142857142857
0.8 35.3055286129971
0.9 44.4533333333333
1 57.6759474091261
};
\addlegendentry{pythia-2.8b  l13}
\addplot [semithick, darkslateblue3654149, mark=*, mark size=1, mark options={solid}]
table {%
0 0
0.1 0
0.2 2.60663507109005
0.3 3.94904458598726
0.4 9.49868073878628
0.5 16.7868852459016
0.6 29.5887662988967
0.7 41.7221135029354
0.8 50.7597801487229
0.9 58.64
1 69.5862335653519
};
\addlegendentry{pythia-6.9b  l7}
\end{axis}

\end{tikzpicture}%
        \vspace{-1cm}
    \end{subfigure}
    \begin{subfigure}[b]{0.45\textwidth}
        \centering
        % This file was created with tikzplotlib v0.10.1.
\begin{tikzpicture}[scale=0.8]

\tikzstyle{every node}=[font=\small]

\definecolor{coral25111684}{RGB}{251,116,84}
\definecolor{darkgray176}{RGB}{176,176,176}
\definecolor{darkslateblue3654149}{RGB}{36,54,149}
\definecolor{firebrick2072831}{RGB}{207,28,31}
\definecolor{lightgray204}{RGB}{204,204,204}
\definecolor{lightgreen151214184}{RGB}{151,214,184}
\definecolor{lightseagreen36152192}{RGB}{36,152,192}
\definecolor{mediumturquoise81188193}{RGB}{81,188,193}
\definecolor{palegoldenrod214239178}{RGB}{214,239,178}
\definecolor{peachpuff252201180}{RGB}{252,201,180}
\definecolor{steelblue33100171}{RGB}{33,100,171}

\begin{axis}[
legend cell align={left},
legend style={
  fill opacity=0.8,
  draw opacity=1,
  text opacity=1,
  at={(0.03,0.97)},
  anchor=north west,
  draw=lightgray204
},
tick align=outside,
tick pos=left,
title={Exact Match per quantile for Phi models},
x grid style={darkgray176},
xlabel={Quantile},
xmajorgrids,
xmin=-0.05, xmax=1.05,
xtick style={color=black},
y grid style={darkgray176},
ylabel={Exact Match (\%)},
ymajorgrids,
ymin=-3.47931167826759, ymax=73.0655452436195,
ytick style={color=black}
]
\addplot [semithick, coral25111684, mark=*, mark size=1, mark options={solid}]
table {%
0 0
0.1 0
0.2 1.18483412322275
0.3 2.92993630573248
0.4 6.06860158311346
0.5 12.2622950819672
0.6 24.5737211634905
0.7 36.0469667318982
0.8 45.4898157129001
0.9 54.5066666666667
1 66.4733178654292
};
\addlegendentry{phi-2  l22}
\addplot [semithick, peachpuff252201180, mark=*, mark size=1, mark options={solid}]
table {%
0 0
0.1 0
0.2 0.710900473933649
0.3 1.01910828025478
0.4 2.55057167985928
0.5 4.32786885245902
0.6 9.57873620862588
0.7 15.6164383561644
0.8 25.2505657937278
0.9 35.9733333333333
1 53.2482598607889
};
\addlegendentry{phi-1.5  l13}
\addplot [semithick, firebrick2072831, mark=*, mark size=1, mark options={solid}]
table {%
0 0
0.1 1.81818181818182
0.2 1.8957345971564
0.3 3.94904458598726
0.4 10.1143359718558
0.5 17.3770491803279
0.6 30.9428284854564
0.7 42.9354207436399
0.8 51.1800840607824
0.9 58.48
1 69.2575406032483
};
\addlegendentry{phi-3  l20}
\end{axis}

\end{tikzpicture}%
        \vspace{-1cm}
    \end{subfigure}
    \caption{Performance on mention generation without context depending on quantiles of entity frequency in the Pile.}
    \vspace{-1cm}
    \label{fig:EMFreqCorr}
\end{figure}
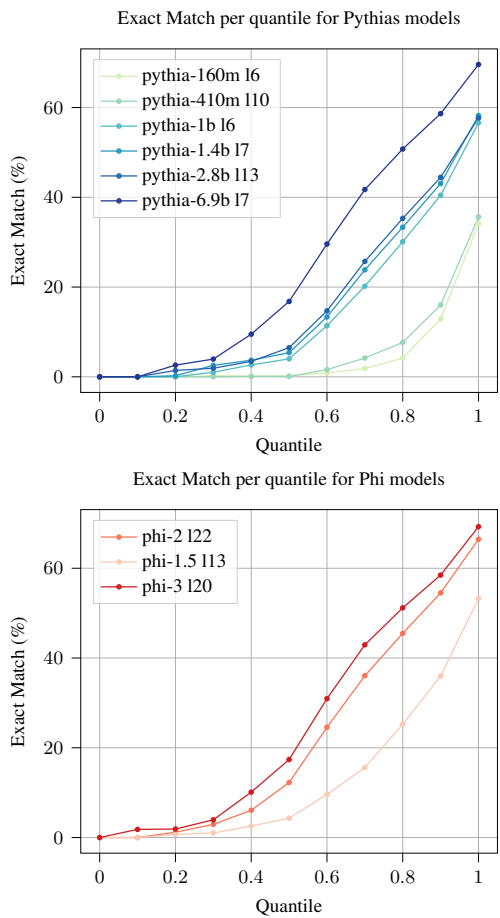

\begin{figure}
    \begin{subfigure}[b]{0.45\textwidth}
        \centering
        % This file was created with tikzplotlib v0.10.1.
\begin{tikzpicture}

\tikzstyle{every node}=[font=\tiny]
\definecolor{coral25111684}{RGB}{251,116,84}
\definecolor{darkgray176}{RGB}{176,176,176}
\definecolor{lightgray204}{RGB}{204,204,204}
\definecolor{mediumblue}{RGB}{0, 51, 204}
\definecolor{mediumseagreen}{RGB}{51, 190, 51}

\begin{axis}[
legend cell align={left},
legend style={
  fill opacity=0.8,
  draw opacity=1,
  text opacity=1,
  at={(0.5,0.09)},
  anchor=south,
  draw=lightgray204
},
width = 8cm,    % Set the desired width
height= 5cm,    % Set the desired heigh
tick align=outside,
tick pos=left,
x grid style={darkgray176},
xlabel={Layer},
xmajorgrids,
xmin=-1.55, xmax=32.55,
xtick style={color=black},
y grid style={darkgray176},
ylabel={Chr-F},
ymajorgrids,
ymin=-0.0452032635318224, ymax=0.94926853416827,
ytick style={color=black}
]
\addplot [semithick, coral25111684, mark=*, mark size=1, mark options={solid}]
table {%
0 0.758602629140516
1 0.810794471603583
2 0.858120774888554
3 0.853357997001868
4 0.853742162997738
5 0.855172205775912
6 0.855706681857492
7 0.852242256115966
8 0.860431670096658
9 0.850281067169087
10 0.856555138292615
11 0.871307473801774
12 0.849364568986936
13 0.85921969651396
14 0.841995370600578
15 0.840156038314798
16 0.786893959470403
17 0.809885227834624
18 0.836147658741361
19 0.796943433148305
20 0.799569521075601
21 0.820880125208741
22 0.802271797402843
23 0.830933901700453
24 0.826659522501509
25 0.829692001091573
26 0.851454178125125
27 0.817489374403807
28 0.817456392223401
29 0.841904841457862
30 0.830137938067141
31 0
};
\addlegendentry{Last}

\addplot [semithick, red, mark=*, mark size=1, mark options={solid}]
table {%
0 0.239541243657802
1 0.289479329516086
2 0.33108238011313
3 0.418981545446277
4 0.39837912436149
5 0.667523212941866
6 0.551308492581334
7 0.698105643125236
8 0.707379432495833
9 0.672710460020404
10 0.600933968034005
11 0.666213428056389
12 0.756740901247845
13 0.791463786732821
14 0.776280154569848
15 0.742921530585909
16 0.735775203525493
17 0.632858915638513
18 0.775808448763294
19 0.592656820441171
20 0.771635760822181
21 0.783392711451221
22 0.705281508963494
23 0.788198535481927
24 0.768165160484625
25 0.810785996920775
26 0.788458015052065
27 0.830830809565228
28 0.7612262043962
29 0.796224844247482
30 0.740037139586046
31 0.788720725850565
};
\addlegendentry{Average}
\addplot [semithick, mediumseagreen, mark=*, mark size=1, mark options={solid}]
table {%
0 0.13491679210487
1 0.197108273440676
2 0.312847187368878
3 0.386682492416786
4 0.461029580313027
5 0.538393799840116
6 0.56576431132459
7 0.492313320421048
8 0.410512306031744
9 0.657745198128175
10 0.591760139789927
11 0.478146417686411
12 0.519920634945042
13 0.473464278178485
14 0.597579486575518
15 0.500648136759746
16 0.481578581050571
17 0.622843258355884
18 0.661979864337571
19 0.609437404119332
20 0.564941758052305
21 0.626824362137934
22 0.504262686527983
23 0.528934938768322
24 0.495953057399217
25 0.469799846126197
26 0.311348256669966
27 0.485088587809382
28 0.48884067779315
29 0.381025526032422
30 0.221938387184167
31 0.372127079164674
};
\addlegendentry{Last + linear}
\addplot [semithick, mediumblue, mark=*, mark size=1, mark options={solid}]
table {%
0 0.360579384349011
1 0.568956235233038
2 0.602363666019493
3 0.873862671055156
4 0.805713798426075
5 0.858238029073542
6 0.897262214407406
7 0.904065270636448
8 0.792419660343756
9 0.822206610015635
10 0.868037793919191
11 0.682035218985134
12 0.702523688990539
13 0.628031245013496
14 0.836262204186986
15 0.864352245147277
16 0.740725966812914
17 0.724665945166289
18 0.675871468195559
19 0.743400264207407
20 0.738147605502125
21 0.75374271351781
22 0.597313188177541
23 0.648398198563603
24 0.65891975828486
25 0.689365634907732
26 0.65254098495818
27 0.566883772715918
28 0.581683870058601
29 0.384431466315627
30 0.264994702353021
31 0.462245637835595
};
\addlegendentry{Average + linear}
\end{axis}

\end{tikzpicture}%
        \vspace{-1cm}
        \subcaption{Performance on \texttt{star\_constellation}}
    \end{subfigure}
    \begin{subfigure}[b]{0.45\textwidth}
        \centering
        % This file was created with tikzplotlib v0.10.1.
\begin{tikzpicture}

\tikzstyle{every node}=[font=\tiny]
\definecolor{coral25111684}{RGB}{251,116,84}
\definecolor{darkgray176}{RGB}{176,176,176}
\definecolor{lightgray204}{RGB}{204,204,204}
\definecolor{mediumblue}{RGB}{0, 51, 204}
\definecolor{mediumseagreen}{RGB}{51, 190, 51}

\begin{axis}[
legend cell align={left},
legend style={
  fill opacity=0.8,
  draw opacity=1,
  text opacity=1,
  at={(0.97,0.03)},
  anchor=south east,
  draw=lightgray204
},
width = 8cm,    % Set the desired width
height= 5cm,    % Set the desired heigh
tick align=outside,
tick pos=left,
x grid style={darkgray176},
xlabel={Layer},
xmajorgrids,
xmin=-1.55, xmax=32.55,
xtick style={color=black},
y grid style={darkgray176},
ylabel={Chr-F},
ymajorgrids,
ymin=0.38729896243963, ymax=0.920114591848388,
ytick style={color=black}
]
\addplot [semithick, red, mark=*, mark size=1, mark options={solid}]
table {%
0 0.702239423152846
1 0.73226323948002
2 0.722266511131598
3 0.751251525129396
4 0.759668190807983
5 0.766817513375878
6 0.768794353547018
7 0.813020588327214
8 0.816174876446899
9 0.801485395972495
10 0.828947979146952
11 0.824634705622888
12 0.840662064224222
13 0.830472041379437
14 0.84049815143197
15 0.826802008349365
16 0.833483197993746
17 0.817364813564434
18 0.832618652938347
19 0.825549819131342
20 0.834787465160054
21 0.828259711635637
22 0.83212986551263
23 0.82786561010142
24 0.831972920365612
25 0.836602301013826
26 0.833198731640498
27 0.837221181215185
28 0.832016385340167
29 0.868168797373356
30 0.839817313331105
31 0.841528227276947
};
\addlegendentry{Average}
\addplot [semithick, coral25111684, mark=*, mark size=1, mark options={solid}]
table {%
0 0.682768906564373
1 0.707577609488511
2 0.722710484086432
3 0.740223528052855
4 0.781639548188938
5 0.784513241867093
6 0.775329215312785
7 0.811315920922902
8 0.820384417697217
9 0.800353955193866
10 0.823166327278495
11 0.832372768705883
12 0.847317477880303
13 0.832364044359536
14 0.849282129305122
15 0.829672617625928
16 0.84571885990087
17 0.833090069136141
18 0.843792395197903
19 0.847053854455676
20 0.845501908450468
21 0.835138032893317
22 0.836444138064791
23 0.848060312383678
24 0.847789509494802
25 0.863647192804721
26 0.868302338144136
27 0.863617761412166
28 0.858566995425594
29 0.885167007927998
30 0.880619909715735
31 0.871194883205954
};
\addlegendentry{Last}
\addplot [semithick, mediumseagreen, mark=*, mark size=1, mark options={solid}]
table {%
0 0.411517854685483
1 0.43339683994018
2 0.507606107145846
3 0.567735048146813
4 0.763578225803037
5 0.787226661214251
6 0.79757712574887
7 0.794711589944103
8 0.821119557453506
9 0.826147249186544
10 0.788973690508732
11 0.822317698942637
12 0.856931442509155
13 0.752885875724907
14 0.838880124591882
15 0.763858198766986
16 0.601402504403082
17 0.802320881217192
18 0.824283921852551
19 0.806131529756284
20 0.873600670148752
21 0.853167744767633
22 0.792676899214775
23 0.785853937961957
24 0.87720751384156
25 0.880816868008052
26 0.841818585246629
27 0.870705232526038
28 0.822021365734378
29 0.895895699602535
30 0.891142942146651
31 0.746101673341209
};
\addlegendentry{Last + linear}
\addplot [semithick, mediumblue, mark=*, mark size=1, mark options={solid}]
table {%
0 0.735033415542004
1 0.749406774708506
2 0.762170981704942
3 0.779237312678903
4 0.789138659400057
5 0.795321033398065
6 0.807012783483107
7 0.817905554298155
8 0.816177009098094
9 0.813249581743648
10 0.815524549389402
11 0.814212551587637
12 0.820830548457989
13 0.815159056523397
14 0.828683830341107
15 0.832782359453743
16 0.822304058374297
17 0.800440366284481
18 0.828297857332577
19 0.834211634492621
20 0.846002559027154
21 0.821337359496938
22 0.828835483895676
23 0.817671630198187
24 0.8445402102938
25 0.811309489548025
26 0.807694624080096
27 0.824806744004301
28 0.781760559784079
29 0.837847625723722
30 0.839919468680283
31 0.832169186341015
};
\addlegendentry{Average + linear}
\end{axis}

\end{tikzpicture}%
        \vspace{-1cm}
        \subcaption{Performance on \texttt{person\_native\_language}}
    \end{subfigure}
    \caption{Chr-F performance on other datasets from \citet{hernandezLinearityRelationDecoding2024}.}
    %\vspace{-0.5cm}
    \label{fig:OtherRelExtractionXPs}
\end{figure}

\subsection{Entity Lens visualizations}\label{sec:EntityLensViz}
\begin{figure*}[t]
    \centering
    \vspace{-.2cm}
    \includegraphics[width=1\linewidth]{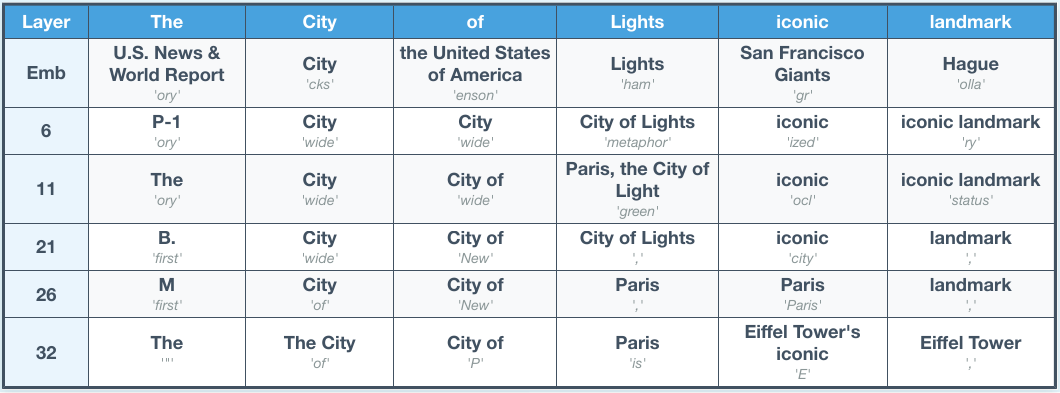}
    \caption{
        The \textit{Entity Lens}, applied using only one task vector ($\theta_{20}$, trained on representations extracted at layer 20 of \textsc{Phi-2} in the \hyperref[par:UncontextualDecoding]{uncontextual} setup).The model still decodes relevant mentions from representations extracted at different layers, showing the generalizing capabilities of $\theta_{20}$ to representations at any layer. This further backs our claim that entity representations are layer agnostic.}
    \label{fig:EntityLens_NoContext_20}
    \vspace{-.2cm}
\end{figure*}

We provide here two example applications of the \textit{Entity Lens}: \Cref{fig:EntityLens_NoContext} for the \hyperref[par:UncontextualDecoding]{uncontextual} setup and \Cref{fig:EntityLens_Context} in the \hyperref[par:contextualDecoding]{contextual} setup.
For any layer $\ell$ and for each token representation $\bz^\ell_k$, we generate a mention with the layer-specific task Vector $\theta_\ell$. To test generalization capabilities through layers, we try \Cref{fig:EntityLens_NoContext_20} to use the same task vector $\theta_{20}$ for all the layers, empirically validating nice generalization capabilities.

\begin{table*}
\resizebox{\textwidth}{!}{%
\begin{tabular}{c|ccccccccccccc|}
\cline{2-14}
                                   & \textbf{G}                  & \textbf{aston}              & \textbf{Julia}                    & \textbf{and}                                                                                & \textbf{Mand}                   & \textbf{el}                     & \textbf{bro}                    & \textbf{t}                      & \textbf{meet}                                                                         & \textbf{,}                                                                                   & \textbf{the}                                                                                & \textbf{latter}                 & \textbf{tells}                                              \\ \hline
\multicolumn{1}{|c|}{\textit{Emb}} & \multicolumn{1}{c|}{Geston} & \multicolumn{1}{c|}{Gaston} & \multicolumn{1}{c|}{Gaston Julia} & \multicolumn{1}{c|}{Mandelbrot}                                                             & \multicolumn{1}{c|}{Mandel}     & \multicolumn{1}{c|}{Mandelbrot} & \multicolumn{1}{c|}{Mandelbrot} & \multicolumn{1}{c|}{Mandelbrot} & \multicolumn{1}{c|}{Mandelbrot}                                                       & \multicolumn{1}{c|}{Mandelbrot}                                                              & \multicolumn{1}{c|}{Mandelbrot}                                                             & \multicolumn{1}{c|}{Mandelbrot} & Mandelbrot                                                  \\ \hline
\multicolumn{1}{|c|}{$\ell$ 6}     & \multicolumn{1}{c|}{G}      & \multicolumn{1}{c|}{Gaston} & \multicolumn{1}{c|}{Gaston Julia} & \multicolumn{1}{c|}{\begin{tabular}[c]{@{}c@{}}Gaston Julia \\ and Mandelbrot\end{tabular}} & \multicolumn{1}{c|}{Mand}       & \multicolumn{1}{c|}{Mandel}     & \multicolumn{1}{c|}{Mandelbrot} & \multicolumn{1}{c|}{Mandelbrot} & \multicolumn{1}{c|}{Meeting}                                                          & \multicolumn{1}{c|}{\begin{tabular}[c]{@{}c@{}}Gaston Julia \\ and Mandelbrot\end{tabular}}  & \multicolumn{1}{c|}{Mandelbrot tells the latter}                                            & \multicolumn{1}{c|}{Mandelbrot} & \begin{tabular}[c]{@{}c@{}}Mandelbrot \\ tells\end{tabular} \\ \hline
\multicolumn{1}{|c|}{$\ell$ 11}    & \multicolumn{1}{c|}{G}      & \multicolumn{1}{c|}{Gaston} & \multicolumn{1}{c|}{Gaston Julia} & \multicolumn{1}{c|}{Gaston Julia}                                                           & \multicolumn{1}{c|}{Mand}       & \multicolumn{1}{c|}{Mandelbrot} & \multicolumn{1}{c|}{Mandelbro}  & \multicolumn{1}{c|}{Mandelbrot} & \multicolumn{1}{c|}{Meeting}                                                          & \multicolumn{1}{c|}{\begin{tabular}[c]{@{}c@{}}Gaston Julia \\ and Mandelbrot\end{tabular}}  & \multicolumn{1}{c|}{Gaston Julia}                                                           & \multicolumn{1}{c|}{Mandelbrot} & Gaston Julia                                                \\ \hline
\multicolumn{1}{|c|}{$\ell$ 16}    & \multicolumn{1}{c|}{G}      & \multicolumn{1}{c|}{Gaston} & \multicolumn{1}{c|}{Gaston Julia} & \multicolumn{1}{c|}{Mandelbrot}                                                             & \multicolumn{1}{c|}{Mand}       & \multicolumn{1}{c|}{Mandel}     & \multicolumn{1}{c|}{Mandelbro}  & \multicolumn{1}{c|}{Mandelbrot} & \multicolumn{1}{c|}{Meeting}                                                          & \multicolumn{1}{c|}{\begin{tabular}[c]{@{}c@{}}Mandelbrot \\ tells, Mandel\end{tabular}}     & \multicolumn{1}{c|}{Mandelbrot}                                                             & \multicolumn{1}{c|}{Mandelbrot} & \begin{tabular}[c]{@{}c@{}}Mandelbrot \\ tells\end{tabular} \\ \hline
\multicolumn{1}{|c|}{$\ell$ 21}    & \multicolumn{1}{c|}{G}      & \multicolumn{1}{c|}{Gaston} & \multicolumn{1}{c|}{Gaston Julia} & \multicolumn{1}{c|}{Mandelbrot tells}                                                       & \multicolumn{1}{c|}{Mand}       & \multicolumn{1}{c|}{Mandelbrot} & \multicolumn{1}{c|}{Mandelbro}  & \multicolumn{1}{c|}{Mandelbrot} & \multicolumn{1}{c|}{Meeting}                                                          & \multicolumn{1}{c|}{Mandelbrot tells}                                                        & \multicolumn{1}{c|}{\begin{tabular}[c]{@{}c@{}}Mandelbrot tells \\ the latter\end{tabular}} & \multicolumn{1}{c|}{Mandelbrot} & Tell                                                        \\ \hline
\multicolumn{1}{|c|}{$\ell$ 26}    & \multicolumn{1}{c|}{G}      & \multicolumn{1}{c|}{Gaston} & \multicolumn{1}{c|}{Gaston Julia} & \multicolumn{1}{c|}{Mandelbrot}                                                             & \multicolumn{1}{c|}{Mand}       & \multicolumn{1}{c|}{Mandelbrot} & \multicolumn{1}{c|}{Mandelbro}  & \multicolumn{1}{c|}{Mandelbrot} & \multicolumn{1}{c|}{Meets}                                                            & \multicolumn{1}{c|}{\begin{tabular}[c]{@{}c@{}}Mandelbrot \\ tells, the latter\end{tabular}} & \multicolumn{1}{c|}{Mandelbrot}                                                             & \multicolumn{1}{c|}{Mandelbrot} & Tell                                                        \\ \hline
\multicolumn{1}{|c|}{$\ell$ 32}    & \multicolumn{1}{c|}{Gaston} & \multicolumn{1}{c|}{Gaston} & \multicolumn{1}{c|}{Gaston Julia} & \multicolumn{1}{c|}{Mandelbrot}                                                             & \multicolumn{1}{c|}{Mandelbrot} & \multicolumn{1}{c|}{Mandelbrot} & \multicolumn{1}{c|}{Mandelbro}  & \multicolumn{1}{c|}{Mandelbrot} & \multicolumn{1}{c|}{\begin{tabular}[c]{@{}c@{}}Mandelbrot \\ tells meet\end{tabular}} & \multicolumn{1}{c|}{Mandelbrot}                                                              & \multicolumn{1}{c|}{Mandelbrot}                                                             & \multicolumn{1}{c|}{Mandelbrot} & \begin{tabular}[c]{@{}c@{}}Mandelbrot \\ tells\end{tabular} \\ \hline
\end{tabular}%
}
\caption{Example application of the \textit{Entity Lens}, applied with a task Vector trained on representations extracted in \textsc{Phi-2} in the \hyperref[par:contextualDecoding]{contextual} setup. We input the sentence ``\texttt{Gaston Julia and Mandelbrot meet, the latter tells}''. We can notably see that the model does associate ``\texttt{the latter}'' with the right entity.}
\label{fig:EntityLens_Context}
\end{table*}

%%%%%%%%%%%%% Perf_Ntoken Frequency
\subsection{Performance as a function of Entity mention size and frequency}\label{sec:perfNtokenFreq}

\begin{figure*}
    \centering
    \includegraphics[width=\linewidth]{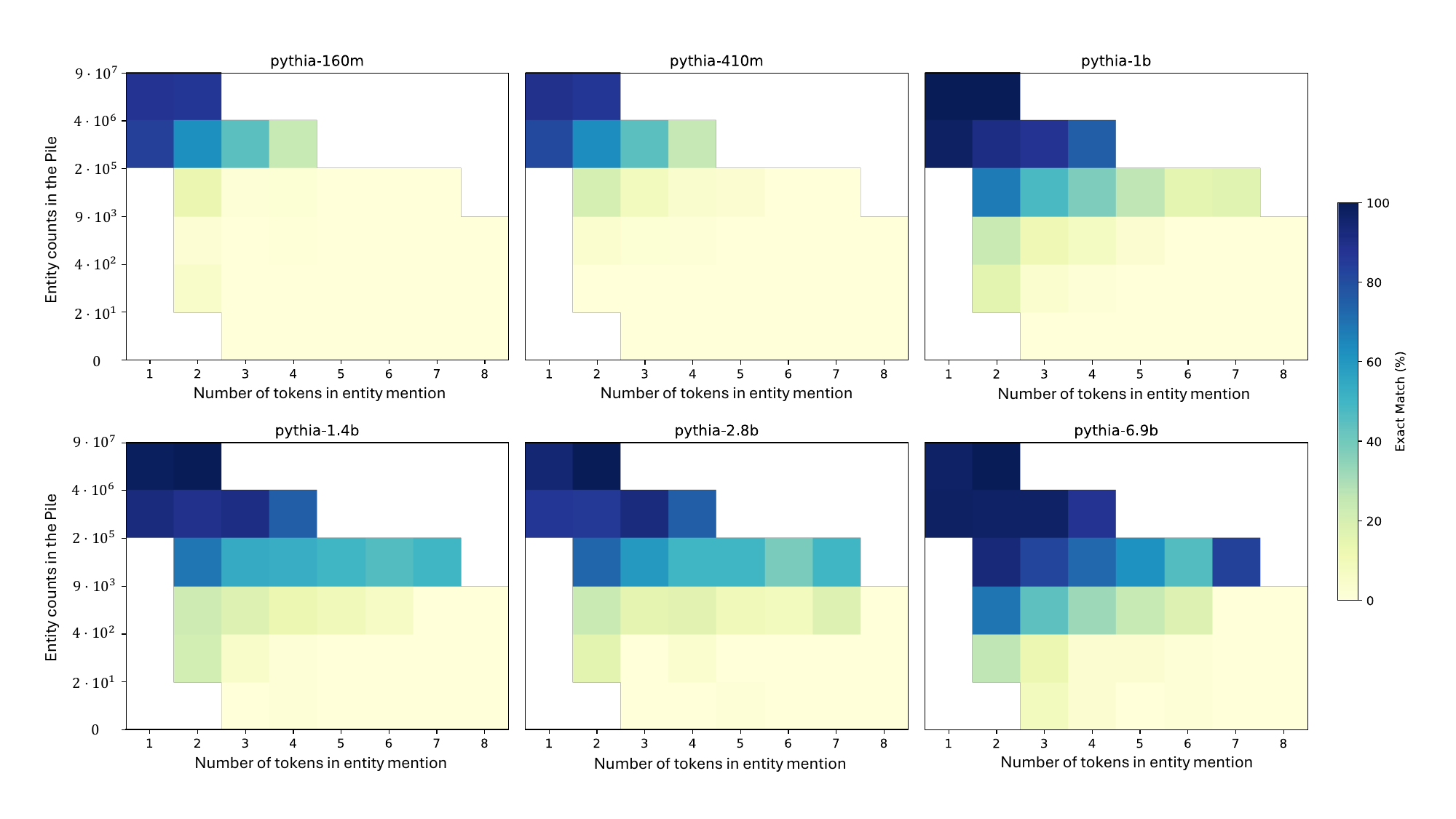}
    \vspace{-.5cm}
    \caption{Reconstruction performance on test set for our \textbf{uncontextual} mention generation experiment. Performance is separated based on the number of tokens that need reconstruction, as well as the n-gram frequency of the mention in \textit{the Pile} \cite{gao2020pile800gbdatasetdiverse}. For each model, we chose the layer with best exact match on the \hyperref[subsec:Dataset]{test set}. Empty cells correspond to count/frequency settings with fewer than 5 samples, making it insufficient to compute performance.}
    \label{fig:PerfNtokenFreq_pythias_noContext}
\end{figure*}

\begin{figure*}
    \centering
    \includegraphics[width=\linewidth]{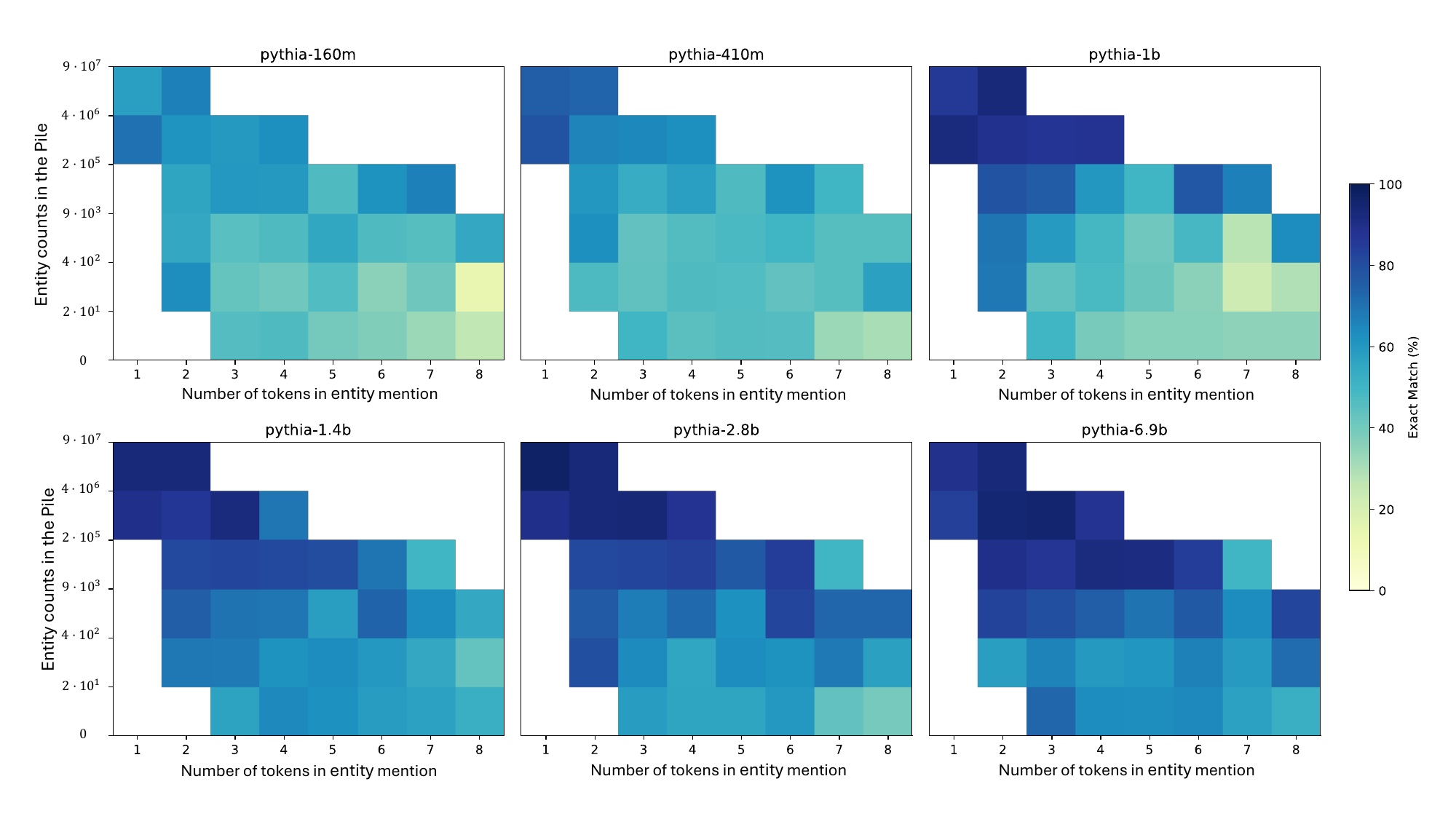}
    \vspace{-.5cm}
    \caption{Reconstruction performance on test set for our \textbf{contextual} mention generation experiment. Performance is separated based on the number of tokens that need reconstruction, as well as the n-gram frequency of the mention in \textit{the Pile} \cite{gao2020pile800gbdatasetdiverse}. For each model, we chose the layer with best exact match on the \hyperref[subsec:Dataset]{test set}. Empty cells correspond to count/frequency settings with fewer than 5 samples, making it insufficient to compute performance.}
    \label{fig:PerfNtokenFreq_pythias_context}
\end{figure*}

\begin{figure*}
    \begin{subfigure}[b]{\textwidth}
        \centering
        \includegraphics[width=\textwidth]{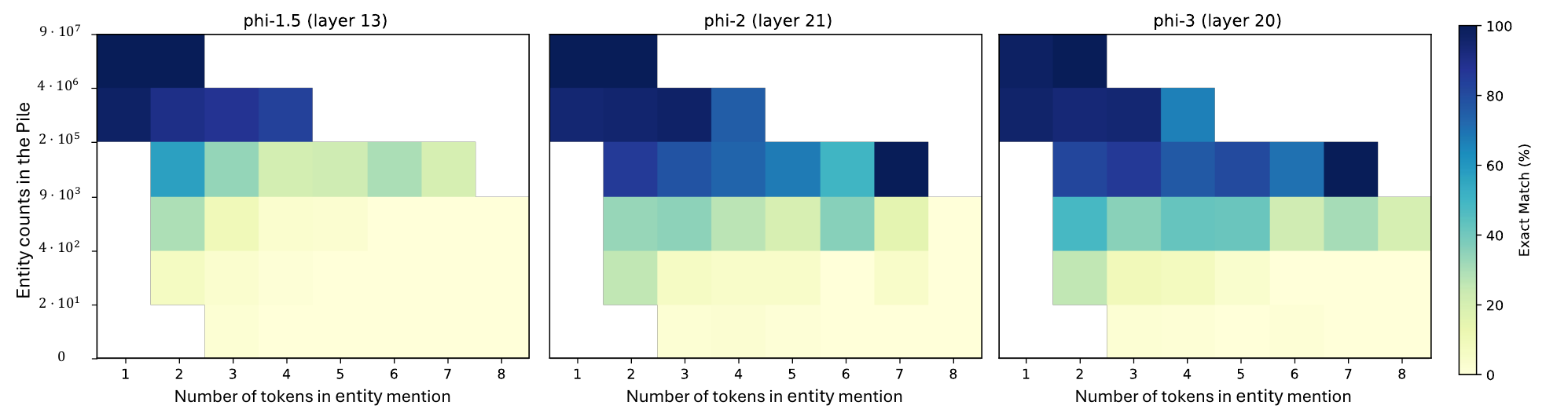}
        \subcaption{Performance analysis for the uncontextual mention generation experiment}
        \label{fig:PerfNtoken_freq_phis_no_context}
    \end{subfigure}
    %\vspace{0cm}
    \begin{subfigure}[b]{\textwidth}
        \centering
        \includegraphics[width=\textwidth]{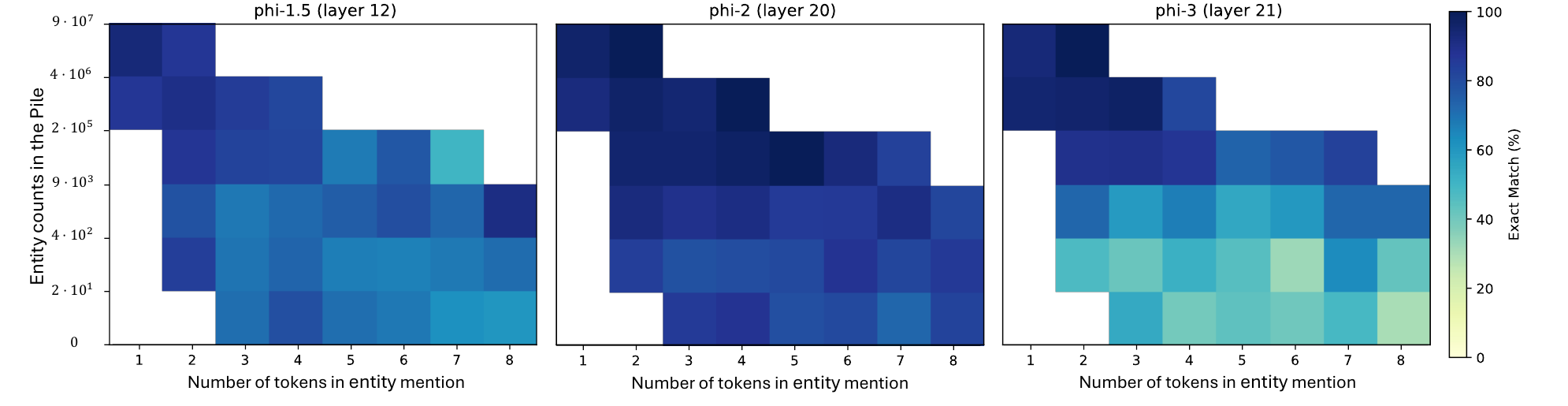}
        \subcaption{Performance analysis for the contextual mention generation experiment}
        \label{fig:PerfNtoken_freq_phis_context}
    \end{subfigure}
    \caption{Reconstruction performance depending on the number of tokens to reconstruct, as well as the n-gram frequency of the entity mention in \textit{the Pile} \cite{gao2020pile800gbdatasetdiverse}. For each model, we chose the layer with best exact match on the \hyperref[subsec:Dataset]{test set}. }
    \label{fig:PerfNtoken_freq_phis}
\end{figure*}

We provide here the complete results for the analysis conducted in \Cref{subsec:Length_frequence}.
Reconstruction performance on test set splitted by mention frequence on \textit{the Pile} and mention length with \textsc{Pythias} models is shown \Cref{fig:PerfNtokenFreq_pythias_context} in the \textit{uncontextual} mention generation setup and  \Cref{fig:PerfNtokenFreq_pythias_noContext} in the \textit{contextual} setup. \Cref{fig:PerfNtoken_freq_phis} gathers the results for the three models from the \textsc{Phi} family.

\end{document}